\RequirePackage{fix-cm}
\documentclass[twocolumn]{svjour3}          % twocolumn
\smartqed  % flush right qed marks, e.g. at end of proof
\usepackage{microtype} %adjusts spaces between words to satsify column width constraints
\usepackage{amsmath,amssymb} % define this before the line numbering.
\usepackage{graphicx}
\usepackage{natbib}
\usepackage{color}
\usepackage[caption=false]{subfig} 
\usepackage[linesnumbered,boxed]{algorithm2e}
\usepackage{url}

\SetAlCapSkip{10pt}  %space caption for algorithm away from box

\begin{document} \sloppy

\title{Do We Need More Training Data?
\thanks{Funding for this research was provided by NSF
IIS-0954083, NSF DBI-1053036, ONR-MURI N00014-10-1-0933, a
Google Research award to CF, and a Microsoft Research gift
to DR.\newline 
The final publication is available at Springer via: \newline
{http://dx.doi.org/10.1007/s11263-015-0812-2}}}

\author{Xiangxin Zhu \and Carl Vondrick \and Charless C. Fowlkes \and Deva Ramanan}
%\authorrunning{Zhu, Vondrick, Fowlkes and Ramanan}

\institute{X. Zhu \and C. Fowlkes \and D. Ramanan \at
  Department of Computer Science, UC Irvine\\
  \email{\{xzhu,fowlkes,dramanan\}@ics.uci.edu} \and
  C. Vondrick \at 
  CSAIL, MIT\\
  \email{vondick@mit.edu}}

%\date{xxx} % The correct dates will be entered by the editor

\maketitle
\begin{abstract}
Datasets for training object recognition systems are
steadily increasing in size.  This paper investigates the
question of whether existing detectors will continue to
improve as data grows, or saturate in performance due to
limited model complexity and the Bayes risk associated with
the feature spaces in which they operate.  We focus on the
popular paradigm of discriminatively trained templates
defined on oriented gradient features.  We investigate the
performance of mixtures of templates as the number of
mixture components and the amount of training data grows.
Surprisingly, even with proper treatment of regularization
and ``outliers'', the performance of classic mixture models
appears to saturate quickly ($\sim$10 templates and
$\sim$100 positive training examples per template).  This is
not a limitation of the feature space as compositional
mixtures that share template parameters via parts and that
can synthesize new templates not encountered during training
yield significantly better performance. Based on our
analysis, we conjecture that the greatest gains in detection
performance will continue to derive from improved
representations and learning algorithms that can make
efficient use of large datasets.

\end{abstract}

\section{Introduction} \label{sec:introduction} 
\begin{figure}[t]
\centering
\includegraphics[width=0.7\columnwidth]{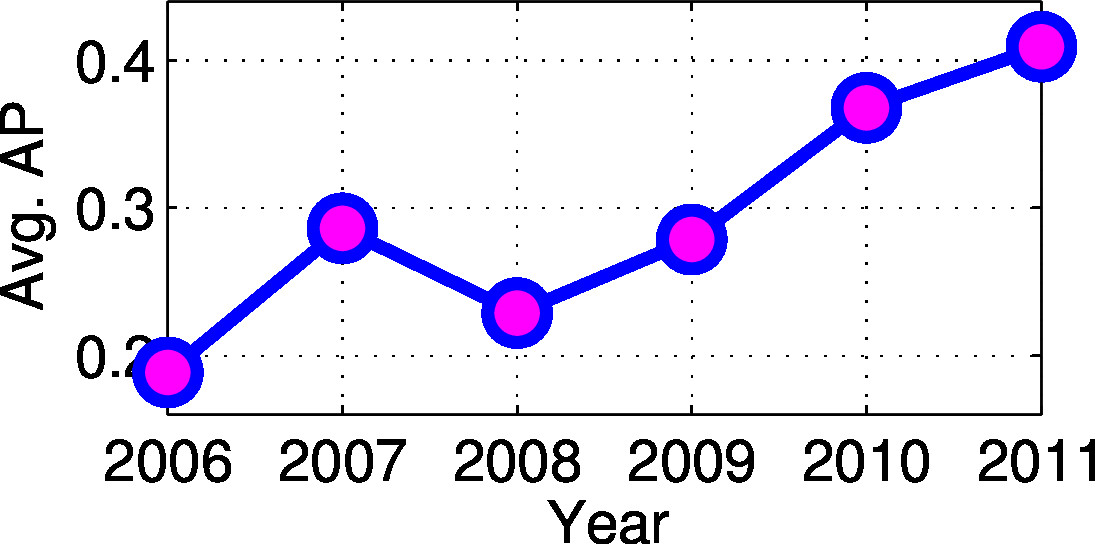}\\
\vspace{10pt}
\includegraphics[width=0.7\columnwidth]{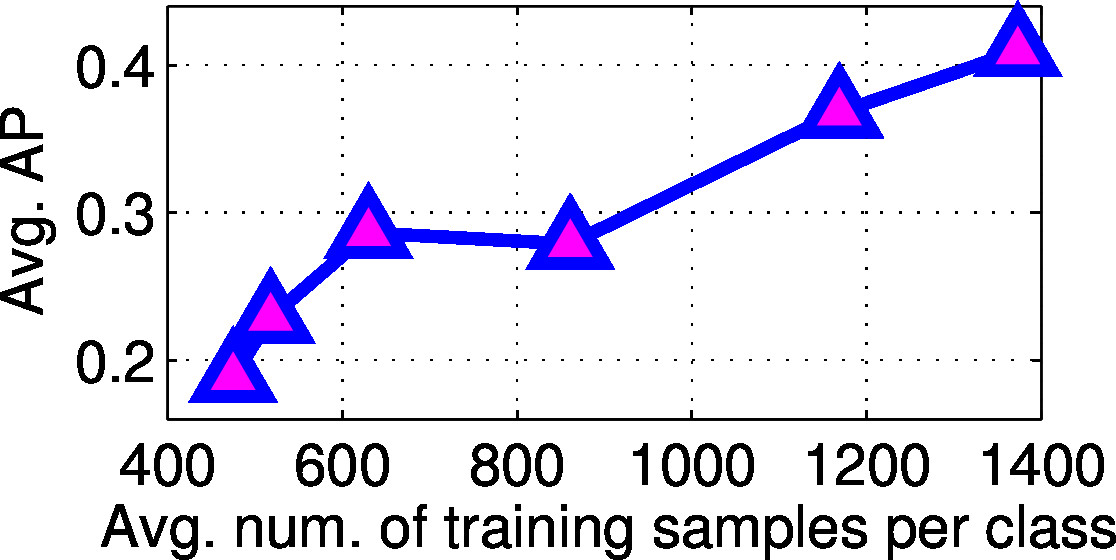}
\caption{The best reported performance on PASCAL VOC
challenge has shown marked increases since 2006 (top). This
could be due to various factors: the dataset itself has
evolved over time, the best-performing methods differ across
years, etc. In the bottom-row, we plot a particular factor
-- training data size -- which appears to correlate well
with performance. This begs the question: has the increase
been largely driven from the availability of larger training
sets?}
\label{fig:voc_best_reported}
\end{figure}

Much of the impressive progress in object detection is built
on the methodologies of statistical machine learning, which
make use of large training datasets to tune model
parameters.  Consider the benchmark results of the
well-known PASCAL VOC object challenge
(Fig.~\ref{fig:voc_best_reported}).  There is a clear trend
of increased benchmark performance over the years as new
methods have been developed.  However, this improvement is
also correlated with increasing amounts of training data.
One might be tempted to simply view this trend as a another
case of the so-called ``effectiveness of big-data'', which
posits that even very complex problems in artificial
intelligence may be solved by simple statistical models
trained on massive datasets~\citep{halevy2009unreasonable}.
This leads us to consider a basic question about the field:
will continually increasing amounts of training data be
sufficient to drive continued progress in object recognition
absent the development of more complex object detection
models?

To tackle this question, we collected a massive training set
that is an order of magnitude larger than existing
collections such as PASCAL \citep{everingham2010pascal}.  We
follow the dominant paradigm of scanning-window templates
trained with linear SVMs on HOG features
\citep{Dalal:CVPR05,felzenszwalb2010object,BourdevMalikICCV09,malisiewicz2011ensemble}, and evaluate detection performance as a function of the
amount of training data and the model complexity. 

\begin{figure}[t]
\centering
\includegraphics[width=0.8\columnwidth]{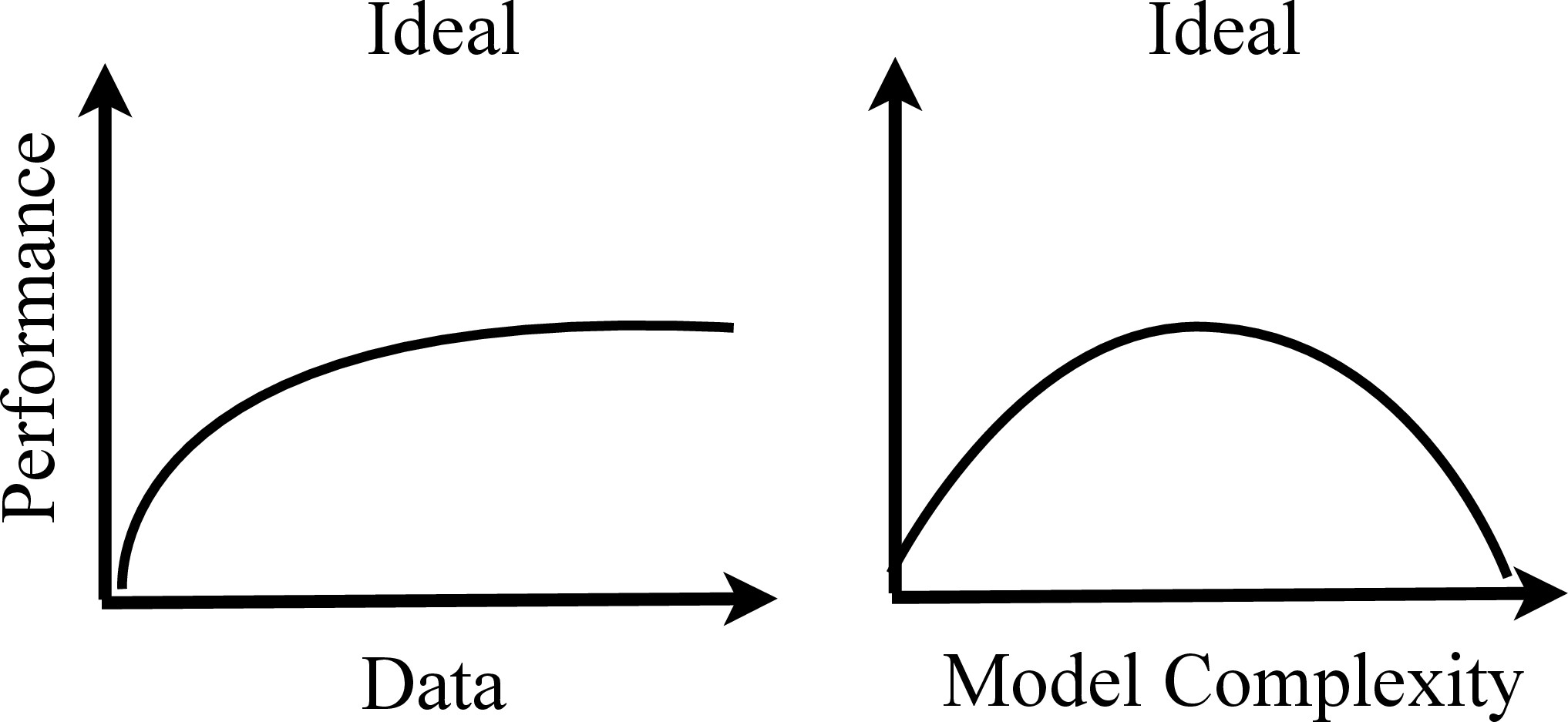}
\caption{We plot idealized curves of performance versus
training dataset size and model complexity. The effect of
additional training examples is diminished as the training
dataset grows ({\bf left}), while we expect performance to
grow with model complexity up to a point, after which an
overly-flexible model overfits the training dataset ({\bf
right}). Both these notions can be made precise with
learning theory bounds, see e.g.
\citep{mcallester1999some}.}
\label{fig:splash}
\end{figure}

{\bf Challenges:} We found there is a surprising amount of
subtlety in scaling up training data sets in current
systems.  For a fixed model, one would expect performance to
generally increase with the amount of data and eventually
saturate (Fig.~\ref{fig:splash}).  Empirically, we often saw
the bizarre result that off-the-shelf implementations show
decreased performance with additional data!  One would also
expect that to take advantage of additional training data,
it is necessary to grow the model complexity, in this case
by adding mixture components to capture different object
sub-categories and viewpoints.  However, even with
non-parametric models that grow with the amount of training
data, we quickly encountered diminishing returns in
performance with only modest amounts of training data. 

We show that the apparent performance ceiling is not a
consequence of HOG+linear classifiers.  We provide an
analysis of the popular deformable part model (DPM), showing
that it can be viewed as an efficient way to implicitly
encode and score an exponentially-large set of rigid mixture
components with shared parameters.  With the appropriate
sharing, DPMs produce substantial performance gains over
standard non-parametric mixture models. However, DPMs have
fixed complexity and still saturate in performance with
current amounts of training data, even when scaled to
mixtures of DPMs. This difficulty is further exacerbated by
the computational demands of non-parametric mixture models,
which can be impractical for many applications.

{\bf Proposed solutions:} In this paper, we offer
explanations and solutions for many of these difficulties.
First, we found it crucial to set model regularization as a
function of training dataset using cross-validation, a
standard technique which is often overlooked in current
object detection systems. Second, existing strategies for
discovering sub-category structure, such as clustering
aspect ratios \citep{felzenszwalb2010object}, appearance
features \citep{Divvala11}, and keypoint labels
\citep{BourdevMalikICCV09}  may not suffice.  We found this
was related to the inability of classifiers to deal with
``polluted'' data when mixture labels were improperly
assigned.  Increasing model complexity is thus only useful
when mixture components capture the ``right'' sub-category
structure. 

To efficiently take advantage of additional training data,
we introduce a non-parametric extension of a DPM which we
call an exemplar deformable part model (EDPM). Notably,
EDPMs increase the expressive power of DPMs with only a
negligible increase in computation, making them practically
useful. We provide evidence that suggests that compositional
representations of mixture templates provide an effective
way to help target the ``long-tail'' of object appearances
by sharing local part appearance parameters across
templates.

Extrapolating beyond our experiments, we see the striking
difference between classic mixture models and the
non-parametric compositional model (both mixtures of linear
classifiers operating on the same feature space) as evidence
that the greatest gains in the near future will not be had
with simple models+bigger data, but rather through improved
representations and learning algorithms.

We introduce our large-scale dataset in
Sec.~\ref{sec:datacollection}, describe our non-parametric
mixture models in Sec.~\ref{sec:mixture}, present extensive
experimental results in Sec.~\ref{sec:exp}, and conclude
with a discussion in Sec.~\ref{sec:conclusion} including
related work.

\section{Big Detection Datasets}
\label{sec:datacollection}
Throughout the paper we carry out experiments using two
datasets. We vary the number of positive training examples,
but in all cases keep the number of negative training images
fixed. We found that performance was relatively static with
respect to the amount of negative training data, once a
sufficiently large negative training set was used.

{\bf PASCAL:} Our first dataset is a newly collected data
set that we refer to as PASCAL-10X and describe in detail in
the following section~\footnote{The dataset can be
downloaded from \url{http://vision.ics.uci.edu/datasets/}}.
This dataset covers the 11 PASCAL categories (see
Fig.~\ref{fig:datasetsize}) and includes approximately 10
times as many training examples per category as the standard
training data provided by the PASCAL detection challenge,
allowing us to explore the potential gains of larger numbers
of positive training instances. We evaluate detection
accuracy on the 11 PASCAL categories from the PASCAL 2010
trainval dataset (because test annotations are not public),
which contains 10000+ images.

{\bf Faces:} In addition to examining performance on PASCAL
object categories, we also trained models for face
detection.  We found faces to contain more structured
appearance variation, which often allowed for more easily
interpretable diagnostic experiments.  Face models are
trained using the CMU MultiPIE
dataset\citep{gross2010multi}, a well-known benchmark
dataset of faces spanning multiple viewpoints, illumination
conditions, and expressions. We use up to $900$ faces across
$13$ view points.  Each viewpoint was spaced $15^{\circ}$
apart spanning $180^{\circ}$.  $300$ of the faces are
frontal, while the remaining $600$ are evenly distributed
among the remaining viewpoints. For negatives, we use $1218$
images from the INRIAPerson database \citep{Dalal:CVPR05}.
Detection accuracy of face models are evaluated on the
annotated face in-the-wild (AFW) \citep{zhu2012face}, which
contains images from real-world environments and tend to
have cluttered backgrounds with large variations in both
face viewpoint and appearance.

\subsection{Collecting PASCAL-10X}

\begin{table}[t!]
\centering
\begin{tabular}{l r r r r}
    & \multicolumn{2}{c}{PASCAL 2010} & \multicolumn{2}{c}{Our Data Set} \\
    Category & Images & Objects & Images & Objects \\
    \hline
    Bicycle      & 471   & 614   & 5,027  & 7,401  \\
    Bus          & 353   & 498   & 3,405  & 4,919  \\
    Cat          & 1,005 & 1,132 & 12,204 & 13,998 \\
    Cow          & 248   & 464   & 3,194  & 6,909  \\
    Dining Table & 415   & 468   & 3,905  & 5,651  \\
    Horse        & 425   & 621   & 4,086  & 6,488  \\
    Motorbike    & 453   & 611   & 5,674  & 8,666  \\
    Sheep        & 290   & 701   & 2,351  & 6,018  \\
    Sofa         & 406   & 451   & 4,018  & 5,569  \\
    Train        & 453   & 524   & 6,403  & 7,648  \\
    TV Monitor   & 490   & 683   & 5,053  & 7,808  \\
    \hline
    Totals       & 4,609 & 6,167 & 50,772 & 81,075 \\
\end{tabular}
\caption{PASCAL 2010 trainval and our data set for select categories. Our data
set is an order of magnitude larger.}
\label{fig:datasetsize}
\end{table}

\begin{table}[t!]
\centering
\begin{tabular}{l r r r}
& & \multicolumn{2}{c}{PASCAL} \\
Attributes & Us     & 2010 & 2007 \\
\hline
Truncated & 30.8 & 31.5 & \textbf{15.8}  \\
Occluded  & 5.9  & 8.6  & 7.1   \\
\hline
Jumping   & 4.0  & 4.3  & \textbf{15.8}  \\
Standing  & 69.9 & 68.8 & \textbf{54.6}  \\
Trotting  & 23.5 & 24.9 & 26.6  \\
Sitting   & 2.0  & 1.4  & 0.7   \\
Other     & 0.0  & 0.5  & 0     \\
\hline
Person Top     & 24.8 & 29.1 & \textbf{57.5}  \\
Person Besides & 8.8 & 10.0 & 8.6   \\ 
No Person      & 66.0 & 59.8 & \textbf{33.8}  \\
\end{tabular}
\caption{Frequencies of attributes (percent) across images
in our 10x horse data set compared to the PASCAL 2010
train-val data set. Bolded entries highlight significant
differences relative to our collected data. Our dataset has
similar attribute distribution to the PASCAL 2010, but
differs significantly from 2007, which has many more
sporting events.}
\label{tab:datasetdistribution}
\end{table}

In this section, we describe our procedure for building a
large, annotated dataset that is as similar as possible to
the PASCAL 2010 for object detection.  We collected images
from Flickr and annotations from Amazon Mechanical Turk
(MTurk), resulting in the data set summarized in
Tab.~\ref{fig:datasetsize}.  We built training sets for 11
of the PASCAL VOC categories that are an order of magnitude
larger than the VOC 2010 standard trainval set.  We selected
these classes as they contain the smallest amount of
training examples, and so are most likely to improve from
additional training data. We took care to ensure
high-quality bounding box annotations and high-similarity to
the PASCAL 2010 dataset.  To our knowledge, this is the
largest publicly available positive training set for these
PASCAL categories.

{\bf Collection:} We downloaded over one hundred thousand
large images from Flickr to build our dataset.  We took care
to directly mimic the collection procedure used by the
PASCAL organizers. We begin with a set of keywords (provided
by the organizers) associated with each object class.  For
each class, we picked a random keyword, chose a random date
since Flickr's launch, selected a random page on the
results, and finally took a random image from that page. We
repeat this procedure until we had downloaded an order of
magnitude larger number of images for each class.

\begin{figure}[t!]
\centering
\includegraphics[width=1\columnwidth]{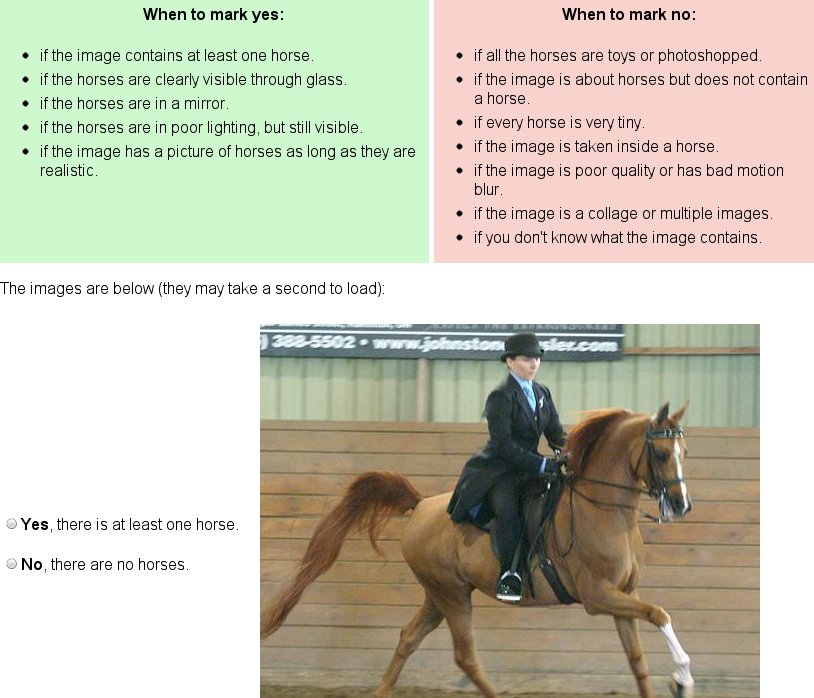}
\includegraphics[width=1\columnwidth]{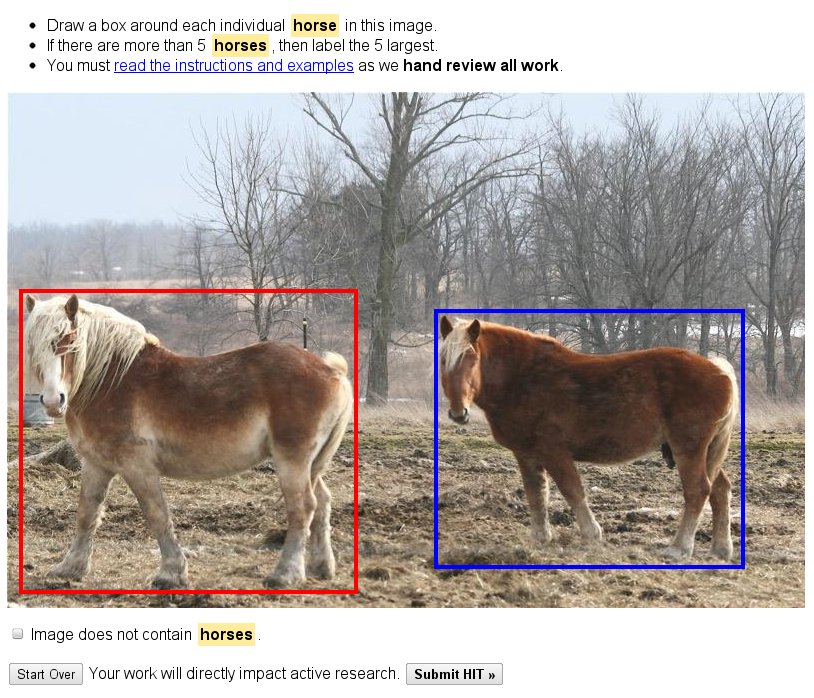}
\caption{Our MTurk user interfaces for image classification
and object annotation. We provided detailed instructions to
workers, resulting in acceptable annotation quality.}
\label{fig:mturkui}
\end{figure}

{\bf Filtering:} The downloaded images from Flickr did not
necessarily contain objects for the category that we were
targeting. We created MTurk tasks that asked workers to
classify the downloaded images on whether they contained the
category of interest. Our user interface in
Fig.~\ref{fig:mturkui} gave workers instructions on how to
handle special cases and this resulted in acceptable
annotation quality without finding agreement between
workers.

{\bf Annotation:} After filtering the images, we created
MTurk tasks instructing workers to draw bounding boxes
around a specific class. Workers were only asked to annotate
up to five objects per image using our interface as in
Fig.\ref{fig:mturkui}, although many workers gave us more
boxes.  On average, our system received annotations at three
images per second, allowing us to build bounding boxes for
10,000 images in under an hour.  As not every object is
labeled, our data set cannot be used to perform detection
benchmarking (it is not possible to distinguish
false-positives from true-negatives). We experimented with
additional validation steps, but found they were not
necessary to obtain high-quality annotations.

\subsection{Data Quality}
\label{sec:datasetdistribution}

To verify the quality of our annotations, we performed an
in-depth diagnostic analysis of a particular category
(horses). Overall, our analysis suggests that our collection
and annotation pipeline produces high-quality training data
that is similar to PASCAL.

{\bf Attribute distribution:} We first compared various
distributions of attributes of bounding boxes from
PASCAL-10X to those from both PASCAL 2010 and 2007 trainval.
Attribute annotations were provided by manual labeling. Our
findings are summarized in
Tab.~\ref{tab:datasetdistribution}. Interestingly, horses
collected in 2010 and 2007 vary significantly, while 2010
and PASCAL-10X match fairly well.  Our images were on
average twice the resolution as those in PASCAL so we scaled
our images down to construct our final dataset.

{\bf User assessment:} We also gauged the quality of our
bounding boxes compared to PASCAL with a user study. We
flashed a pair of horse bounding boxes, one from PASCAL-10X
and one from PASCAL 2010, on a screen and instructed a
subject to label which appeared to be better example. Our
subject preferred the PASCAL 2010 data set 49\% of the time
and our data set 51\% of the time. Since chance is 50\%-50\%
and our subject operated close to chance, this further
suggests PASCAL-10X matched well with PASCAL. Qualitatively,
the biggest difference observed between the two datasets was
that PASCAL-10X bounding boxes tend to be somewhat
``looser'' than the (hand curated) PASCAL 2010 data.

{\bf Redundant annotations:} 
We tested the use of multiple annotations for removing
poorly labeled positive examples. All horse images were
labeled twice, and only those bounding boxes that agreed
across the two annotation sessions were kept for training.
We found that training on these cross-verified annotations
did not significantly affect the performance of the learned
detector.

\begin{figure}[t]
\begin{center}
\subfloat[Unsupervised]{\includegraphics[width=\columnwidth]{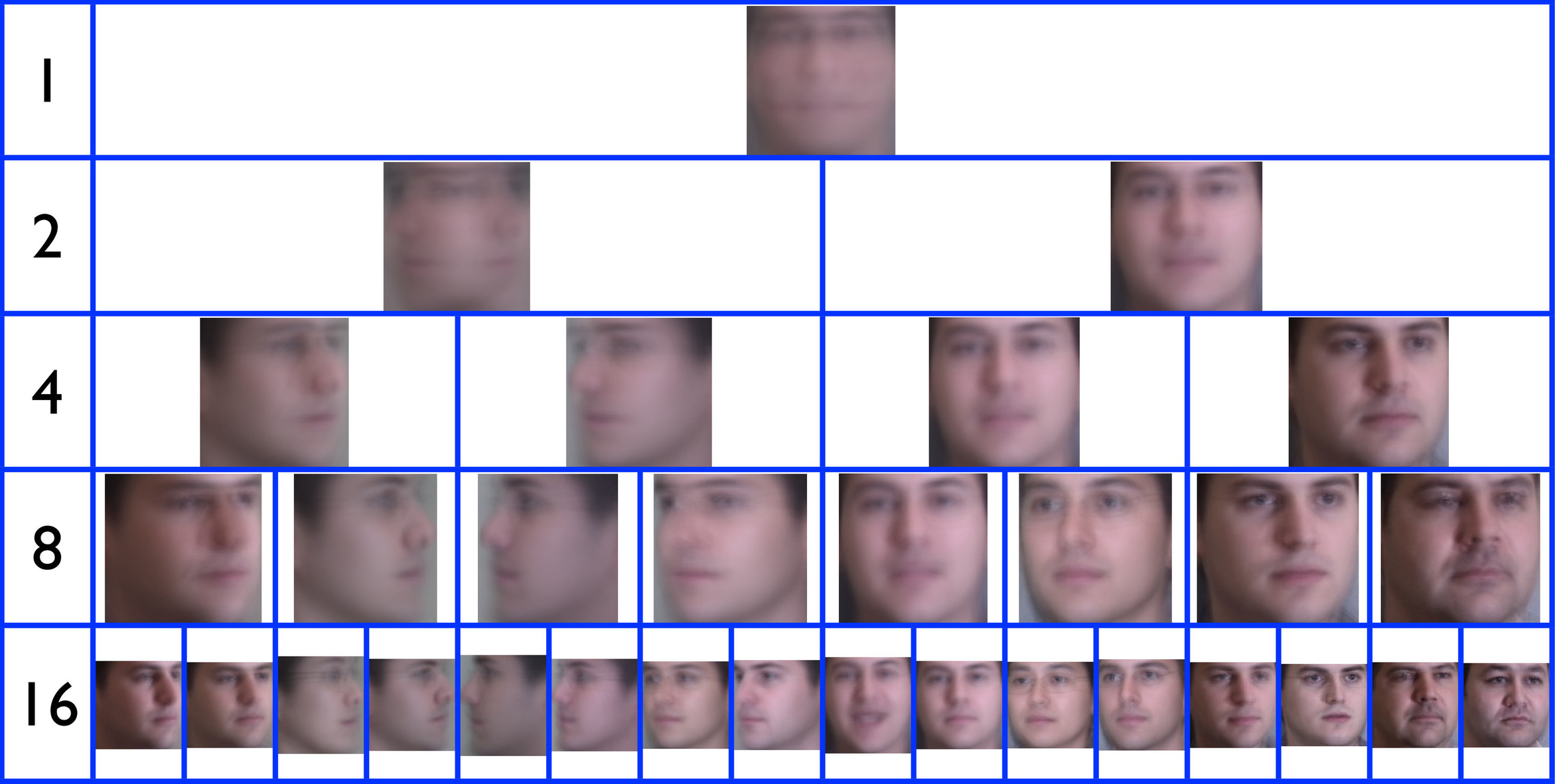}}\\
\subfloat[Supervised]{\includegraphics[width=\columnwidth]{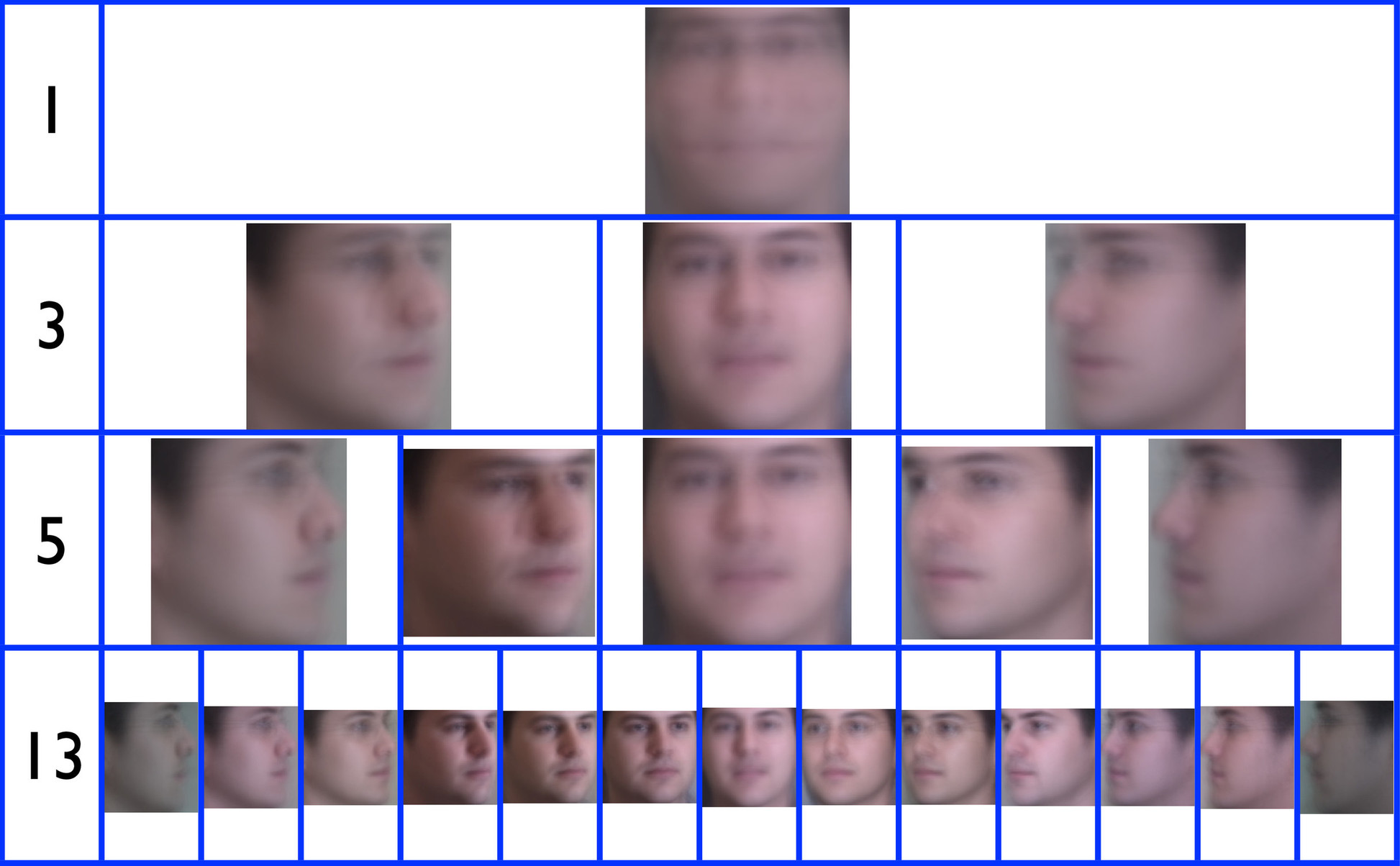}}
\caption{We compare supervised versus automatic (k-means)
approaches for clustering by displaying the average RGB
image of each cluster. The supervised methods use viewpoint
labels to cluster the training data. Because our face data
is relatively clean, both obtain reasonably good clusters.
However, at some levels of the hierarchy, unsupervised
clustering does seem to produce suboptimal partitions - for
example, at $K=2$.  There is no natural way to group
multi-view faces into two groups.  Automatically selecting
$K$ is a key difficulty with unsupervised clustering
algorithms.}
\label{fig:face_cluster}
\end{center}
\end{figure}

\begin{figure}[t!]
\begin{center}
\subfloat[Unsupervised]{\includegraphics[width=\columnwidth]{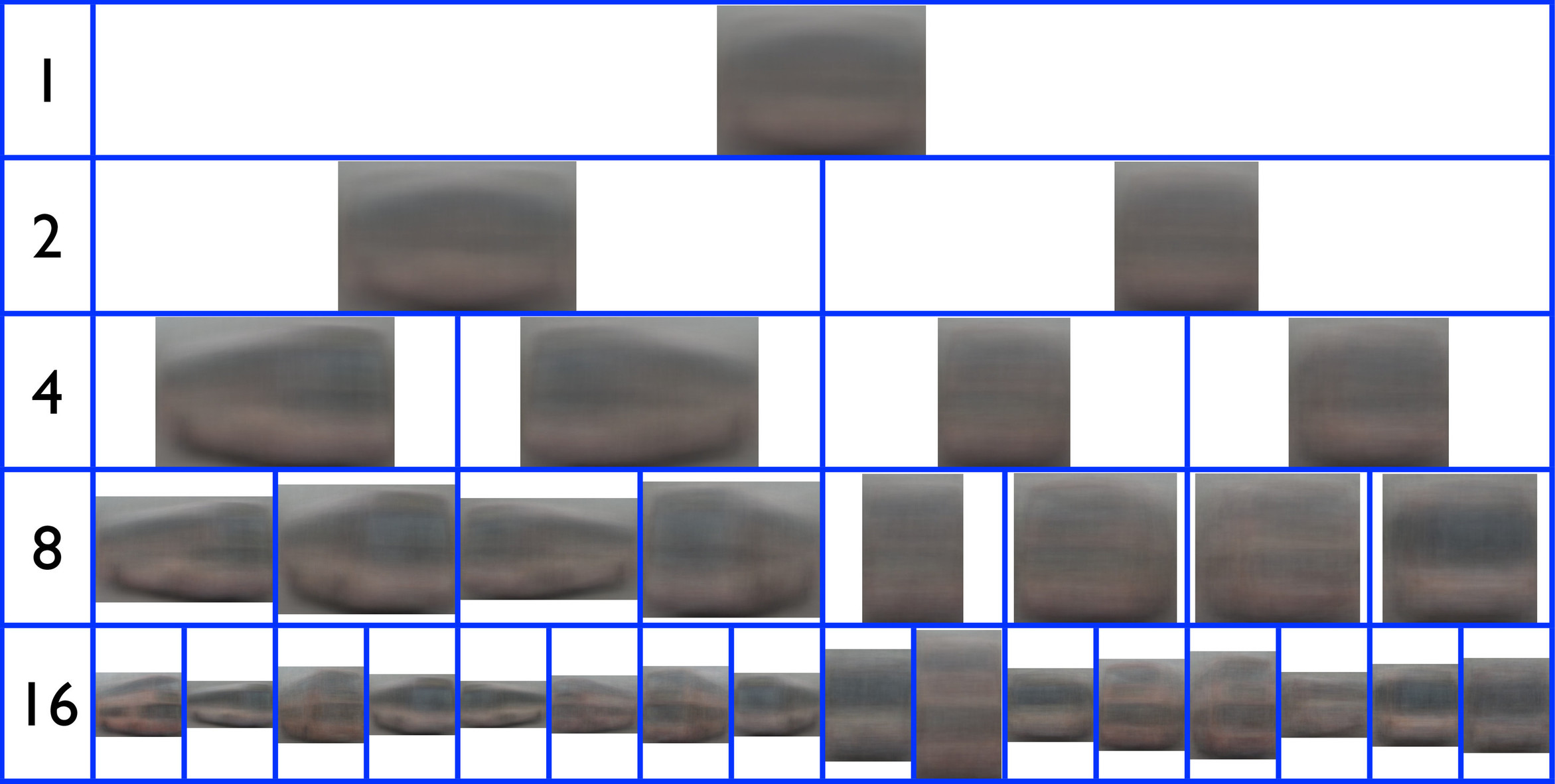}} \\
\subfloat[Supervised]{\includegraphics[width=\columnwidth]{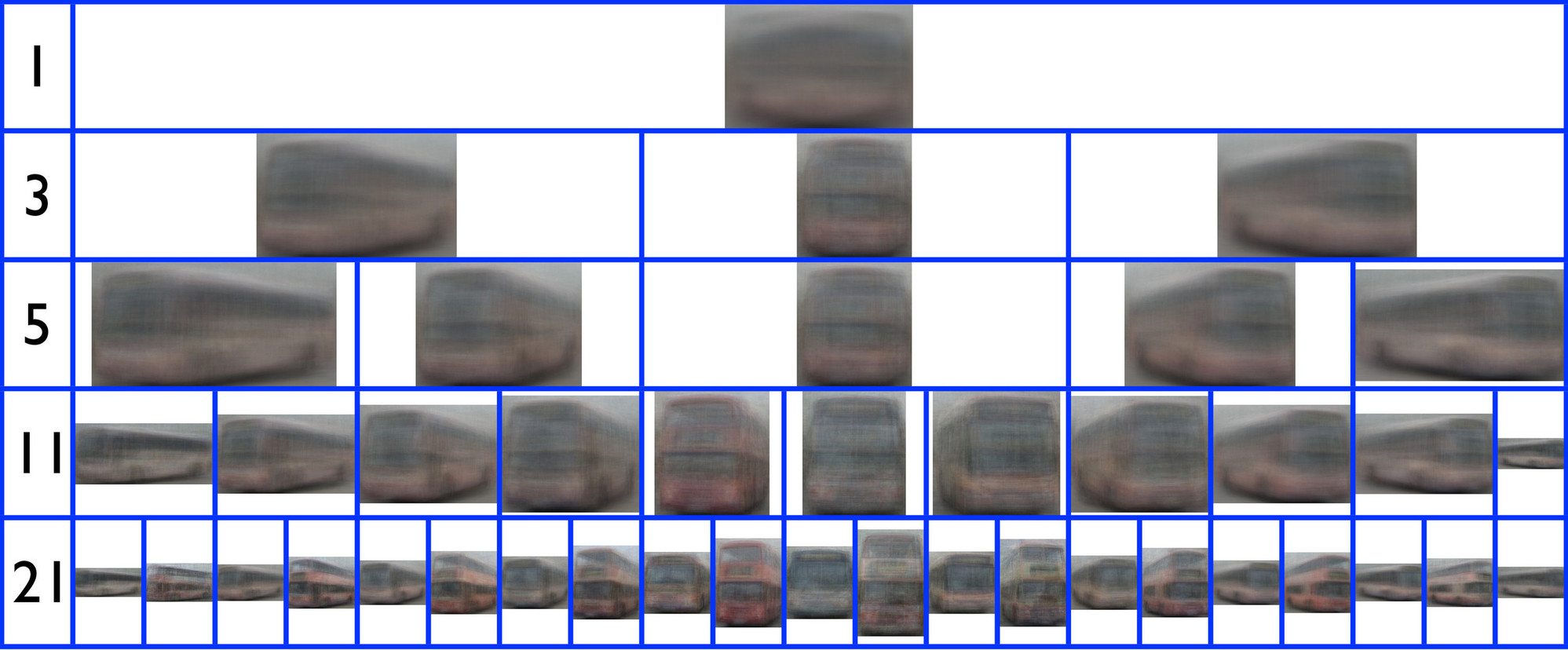}}
\caption{We compare supervised versus automatic (k-means)
approaches for clustering images of PASCAL buses.
Supervised clustering produces more clear clusters, e.g.\
the 21 supervised clusters correspond to viewpoints and
object type (single vs double-decker). Supervised clusters
perform better in practice, as we show in
Fig.~\ref{fig:clean}.}
\label{fig:cluster}
\end{center}
\end{figure}

\section{Mixture models}
\label{sec:mixture}
To take full advantage of additional training data, it is
vital to grow model complexity. We accomplish this by adding
a mixture component to capture additional ``sub-category''
structure. In this section, we describe various approaches
for learning and representing mixture models. Our basic
  building block will be a mixture of linear classifiers, or
  templates. Formally speaking, we compute the detection
  score of an image window $I$ as:
\begin{align}
S(I) = \max_{m} \Big[ w_m \cdot \phi(I) + b_m \Big] \label{eq:mix}
\end{align}
\noindent where $m$ is a discrete mixture variable,
$\Phi(I)$ is a HOG image descriptor \citep{Dalal:CVPR05},
$w_m$ is a linearly-scored template, and $b_m$ is an
(optional) bias parameter that acts as a prior that favors
particular templates over others.

\subsection{Independent mixtures}
In this section, we describe approaches for learning mixture
models by clustering positive examples from our training
set. We train independent linear classifiers $(w_m,b_m)$
using positive examples from each cluster. One difficulty in
evaluating mixture models is that fluctuations in the
(non-convex) clustering results may mask variations in
performance we wish to measure.  We took care to devise a
procedure for varying $K$ (the number of clusters) and $N$
(the amount of training data) in such a manner that would
reduce stochastic effects of random sampling. 

{\bf Unsupervised clustering:} 
For our unsupervised baseline, we cluster the positive
training images of each category into 16 clusters using
hierarchical k-means, recursively splitting each cluster
into $k=2$ subclusters. For example, given a fixed training
set, we would like the cluster partitions for $K=8$ to
respect the cluster partition of $K=4$. To capture both
appearance and shape when clustering, we warp an instance to
a canonical aspect ratio, compute its HOG descriptor (reduce
the dimensionality with PCA for computational efficiency),
and append the aspect ratio to the resulting feature vector.

\begin{algorithm}[h]
{\footnotesize
%\SetLine
\KwIn{$\{N_n\}$; $\{S^{(i)}\}$ }
\KwOut{$\{C_n^{(i)}\}$} 
$C_0^{(i)} = S^{(i)}, \quad C_n^{(i)}=\emptyset \qquad \forall i, \forall n > 1$ \\
\For(\tcp*[f]{For each $N_n$}){$n = 1:end$}  { 
%    $cnt = 0$\\
%\While{$cnt\leq N_n$}{
\For{$t = 1:N_n$}{
   $z \sim \frac{|C_{n-1}^{(z)}|}{\sum_j|C_{n-1}^{(j)}|}$ \tcp*{Pick a cluster randomly}
   $C_n^{(z)} \Leftarrow C_{n-1}^{(z)}$ \tcp*{sample $z$th cluster without replacement}
%    $cnt = cnt+1$\\
    }
  }
}
\caption{
Partitioned sampling of the clusters. $N_n$ is the number of
samples to return for set $n$ with $N_0=N_{max}$;
$N_{n}>N_{n+1}$. $S^{(i)}$ is the $i^{th}$ cluster from the
lowest level of the hierarchy (e.g., with $K=16$ clusters)
computed on the full dataset $N_{max}$. Steps 4-5 randomly
samples $N_n$ training samples from $\{C_{n-1}^{(i)}\}$ to
construct $K$ sub-sampled clusters $\{C_n^{(i)}\}$, each of
which contain a subset of the training data while keeping
the same distribution of the data over clusters.}
\label{alg:partitionedsampling}
\end{algorithm}

{\bf Partitioned sampling:} 
Given a fixed training set of $N_{max}$ positive images, we would like to
construct a smaller sampled subset, say of $N=\frac{N_{max}}{2}$ images, whose
cluster partitions respect those in the full dataset.  This is similar in
spirit to stratified sampling and attempts to reduce variance in our
performance estimates due to ``binning artifacts'' of inconsistent cluster
partitions across re-samplings of the data.

To do this, we first hierarchically-partition the full set of $N_{max}$ images
by recursively applying k-means.  We then subsample the images in the leaf
nodes of the hierarchy in order to generate a smaller hierarchically
partitioned dataset by using the same hierarchical tree defined over the
original leaf clusters.  This sub-sampling procedure can be applied repeatedly
to produce training datasets with fewer and fewer examples that still respects
the original data distribution and clustering.

The sampling algorithm, shown in Alg.~\ref{alg:partitionedsampling}, yields a set of
partitioned training sets, indexed by $(K,N)$ with two properties: (1) for a
fixed number of clusters $K$, each smaller training set is a subset of the
larger ones, and (2) given a fixed training set size $N$, small clusters are
strict refinements of larger clusters.  We compute confidence intervals in our
experiments by repeating this procedure multiple times to resample the dataset
and produce multiple sets of $(K,N)-$consistent partitions.

{\bf Supervised clustering:}
To examine the effect of supervision, we cluster the training data by manually
grouping visually similar samples. For CMU MultiPIE, we define clusters using
viewpoint annotations provided with the dataset. We generate a hierarchical
clustering by having a human operator merge similar viewpoints, following the
partitioned sampling scheme above. Since PASCAL-10X does not have viewpoint
labels, we generate an  ``over-clustering'' with k-means with a large $K$, and
have a human operator manually merge clusters. Fig.~\ref{fig:face_cluster} and
Fig.~\ref{fig:cluster} show example clusters for faces and buses.

\subsection{Compositional mixtures}

In this section, we describe various architectures for compositional mixture
models that share information between mixture components. We share local
spatial regions of templates, or parts. We begin our discussion by reviewing
standard architectures for deformable part models (DPMs), and show how they can
be interpreted and extended as high-capacity mixture models.

{\bf Deformable Part Models (DPMs):}
We begin with an analysis that shows that DPMs are equivalent to an
exponentially-large mixture of rigid templates Eqn.~\eqref{eq:mix}. This allows
us to analyze (both theoretically and empirically) under what conditions a
classic mixture model will approach the behavior of a DPM. Let the location of
part $i$ be $(x_i,y_i)$. Given an image $I$, a DPM scores a configuration of
$P$ parts $(x,y) = \{(x_i,y_i): i = 1..P\}$ as:
\begin{align}
S_{DPM}(I) &= \max_{x,y} S(I,x,y) \quad \text{where}  \nonumber\\
S(I,x,y) &= \sum_{i = 1}^P \sum_{(u,v) \in W_i}
\alpha_i[u,v] \cdot \phi(I,x_i + u,y_i + v) \nonumber\\
&\!+\!\sum_{ij \in E} \beta_{ij} \cdot \psi(x_i\!-\!x_j\!-\!a_{ij}^{(x)}, y_i\!-\!y_j\!-\!a_{ij}^{(y)}) \label{eqn:rmp} 
\end{align}
\noindent where $W_i$ defines the spatial extent (length and width) of
part $i$. The first term defines a local appearance score, where
$\alpha_i$ is the appearance template for part $i$ and
$\phi(I,x_i,y_i)$ is the appearance feature vector extracted from location
$(x_i,y_i)$. The second term defines a pairwise deformation model that
scores the relative placement of a pair of parts with respect to an
anchor position $(a^{(x)}_{ij},a^{(y)}_{ij})$. For simplicity, we have assumed all filters are defined at the same scale, though the above can be extended to the multi-scale case. When the associated
relational graph $G=(V,E)$ is tree-structured, one can compute the
best-scoring part configuration $\max_{(x,y) \in \Omega} S(I,x,y)$
with dynamic programming, where $\Omega$ is the space of possible part
placements. Given that each of $P$ parts can be placed at one of $L$
locations, $|\Omega| = L^P \approx 10^{20}$ for our models.

By defining index variables in image coordinates $u' = x_i + u$ and $v' = y_i + v$, we can rewrite Eqn.~\eqref{eqn:rmp} as: 
\begin{align}
&S(I,x,y) =  \sum_{u',v'} \sum_{i =1}^P \alpha_i[u'-x_i,v'-y_i] \cdot \phi(I,u',v')  \nonumber\\
&\ \quad \quad \quad \quad + \sum_{ij \in E} \beta_{ij}\cdot \psi_{ij}(x_i-x_j - a^{(x)}_{ij},y_i-y_j - a^{(y)}_{ij}) \nonumber\\
&\ \ \ = \Big( \sum_{u',v'} w(x,y)[u',v'] \cdot \phi(I,u',v')  \Big) + b(x,y) \nonumber\\
&\ \ \ = w(x,y) \cdot \phi(I) + b(x,y) \label{eq:epm}
\end{align}
\noindent where $w(x,y)[u',v'] = \sum_{i =1}^P
\alpha_i[u'-x_i,v'-y_i]$. For notational convenience, we assume parts
templates are padded with zeros outside of their default spatial
extent.

From the above expression it is easy to see that the DPM scoring
function is formally equivalent to an exponentially-large mixture model where
each mixture component $m$ is indexed by a particular configuration of parts
$(x,y)$. The template corresponding to each mixture component $w(x,y)$ is
constructed by adding together parts at shifted locations. The bias
corresponding to each mixture component $b(x,y)$ is equivalent to the spatial
deformation score for that configuration of parts. 

DPMs differ from classic mixture models previously defined in that they (1) share parameters across a large
number of mixtures or rigid templates, (2) extrapolate by ``synthesizing'' new
templates not encountered during training, and finally, (3) use dynamic
programming to efficiently search over a large number of templates. 

\begin{figure*}[t!]
\centering
\includegraphics[width=.9\textwidth]{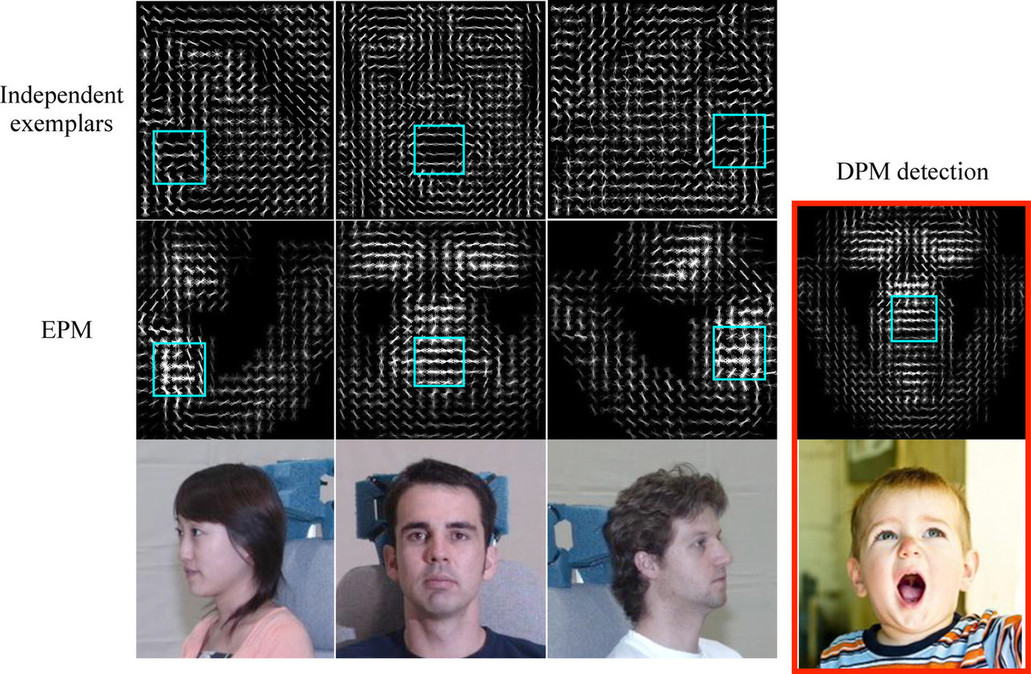}
\caption{Classic exemplars vs EPMs. On the top row, we show three rigid templates trained as independent exemplar mixtures. Below them, we show their counterparts from an exemplar part model (EPM), along with their corresponding training images. EPMs
  share spatially-localized regions (or ``parts'') between
  mixtures. Each rigid mixture is a superposition of overlapping
  parts. A single part is drawn in blue. We show parts on the top row to emphasize that these template regions are trained independently. On the [right], we show a
  template which is implicitly synthesized by a DPM for a
  novel test image on-the-fly. In Fig.~\ref{fig:bayes}, we show that
  both sharing of parameters between mixture components and implicit
  generation of mixture components corresponding to unseen part
  configurations contribute to the strong performance of a
  DPM. 
 \label{fig:EPM}}
\end{figure*}

{\bf Exemplar Part Models (EPMs):} To analyze the relative importance of
part parameter sharing and extrapolation to new part placements, we define a
part model that limits the possible configurations of parts to those seen in
the $N$ training images, written as 
\begin{align}
S_{EPM}(I) = \max_{(x,y) \in \Omega_N} S(I,x,y) \quad \text{where}
\quad \Omega_N \subseteq \Omega.
\end{align}
We call such a model an Exemplar Part Model (EPM), since it can also be
interpreted as set of $N$ rigid exemplars with shared parameters. EPMs are not
to be confused with exemplar DPMs (EDPMs), which we will shortly introduce as
their deformable counterpart. EPMs can be optimized with a discrete enumeration
over $N$ rigid templates rather than dynamic programming. However, by caching
scores of the local parts, this enumeration can be made quite efficient even
for large $N$.  EPMs have the benefit of sharing, but cannot synthesize new
templates that were not present in the training data.  We visualize example EPM
templates in Fig.~\ref{fig:EPM}. 

To take advantage of additional training data, we would like to explore non-parametric mixtures of DPMs. One practical issue is that of computation. We show that with a particular form of sharing, one can construct non-parametric DPMs that are no more computationally complex than standard DPMs or EPMs, but considerably more flexible in that they extrapolate multi-modal shape models to unseen configurations.

{\bf Exemplar DPMs (EDPMs):} To describe our model, we first define a mixture of DPMs
with a shared appearance model, but mixture-specific shape models.
In the extreme case, each mixture will consist of a single training exemplar.
We describe an approach that shares both the part filter computations {\em and}
dynamic programming messages across all mixtures, allowing us to eliminate almost all of
the mixture-dependant computation. Specifically, we consider mixture of
DPMs of the form:

\begin{align}
S(I) = \max_{m \in \{1 \dots M\}} \max_{z \in \Omega} \Big[
w(z) \cdot \phi(I)  + b_m(z) \Big] %\\
\end{align}
where $z=(x,y)$ and we write a DPM as an inner maximization over an
exponentially-large set of templates indexed by $z \in \Omega$, as in
Eqn.~\eqref{eq:epm}. Because the appearance model does not depend on $m$, we can
write:
\begin{align}
S(I) = \max_{z \in \Omega} \Big[ 
w(z) \cdot \phi(I)  + b(z) \Big] \label{eq:all}
\end{align}
where $b(z) = \max_m b_m(z)$. Interestingly,
we can write the DPM, EPM, and EDPM in the form of Eqn.~\eqref{eq:all} by 
simply changing the shape model $b(z)$:

\begin{align}
b_{DPM}(z) &= \sum_{ij \in E} \beta_{ij} \cdot \psi(z_i - z_j -
a_{ij})\\
b_{EDPM}(z) &= \max_{m \in \{1 \ldots M\}} \sum_{ij \in E} \beta_{ij} \cdot \psi(z_i - z_j -
a_{ij}^m) \label{eq:edpm}\\
b_{EPM}(z) &= b_{DPM}(z) + b_{EDPM}^*(z)
%\label{eqn:exDPM}
\end{align}
\noindent where $a_{ij}^m$ is the anchor position for part $i$ and $j$ in
mixture $m$. We write $b_{EDPM}^*(z)$ to denote a limiting case of $b_{EDPM}(z)$
with $\beta_{ij} = -\infty$ and thus takes on a value of $0$ when $z$ has the
same relative part locations as some exemplar $m$ and $-\infty$ otherwise.

\begin{figure}[t!]
\centering
\includegraphics[width=\columnwidth]{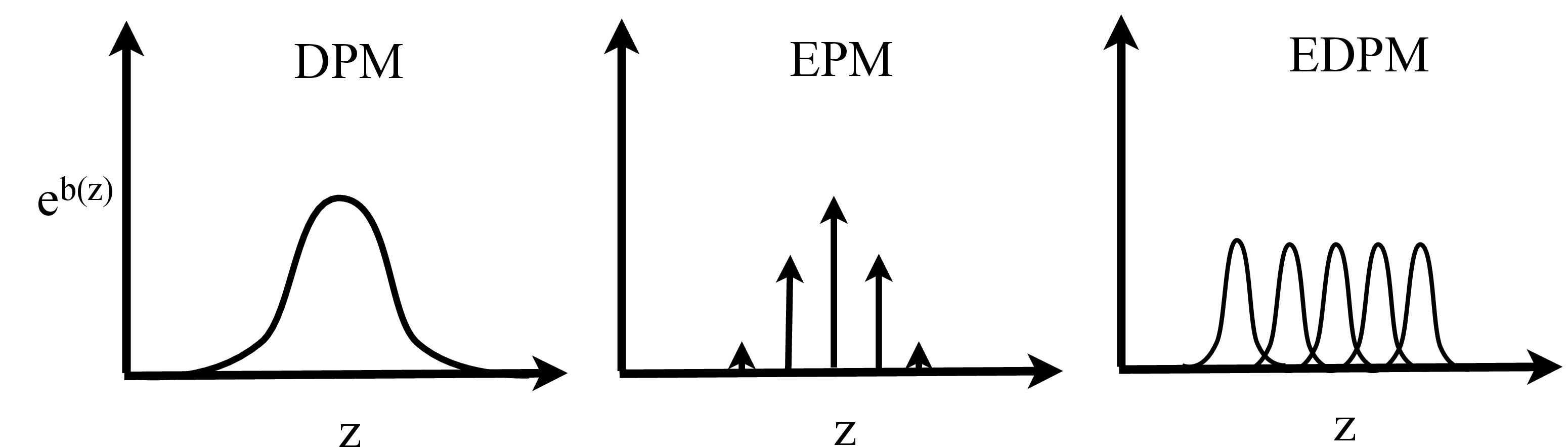}
\caption{We visualize exponentiated shape models $e^{b(z)}$
  corresponding to different part models. A DPM uses a
  unimodal Gaussian-like model ({\bf left}), while a EPM allows for only a
  discrete set of shape configurations encountered at training ({\bf
    middle}). An EDPM non-parametrically models an arbitrary shape
  function using a small set of basis functions. From this
  perspective, one can view EPMs as special cases of EDPMs using
  scaled delta functions as basis functions.}
\label{fig:shape}
\end{figure}

While the EPM only considers $M$ different part configurations to occur at test
time, the EDPM extrapolates away from these shape exemplars.  The spring
parameters $\beta$ in the EDPM thus play a role similar to the kernel width 
in kernel density estimation.  We show a visualization of these shape
models as probabilistic priors in Fig.~\ref{fig:shape}.

{\bf Inference:} We now show that inference on EDPMs (Eqn.~\ref{eq:edpm}) can be quite efficient. Specifically, inference on a star-structured EDPM is no more expensive than a EPM built from the same training
examples. Recall that EPMs can be efficiently optimized with a discrete
enumeration of $N$ rigid templates with ``intelligent caching'' of
part scores. % can be evaluated by caching the
%local part scores
Intuitively, one computes a response map for each part, and then scores a rigid
template by looking up shifted locations in the response
maps. EDPMs operate in a similar same manner, but one
convolves a ``min-filter'' with each response map before
looking up shifted locations.  To be precise, we explicitly
write out the message-passing equations for a
star-structured EDPM below, where we assume part $i=1$ is
the root without loss of generality:
%\begin{align}
\begin{equation}
S_{EDPM}(I) = \max_{z_1,m} \Big[ \alpha_1 \cdot \phi(I,z_1) + \sum_{j > 1} m_j(z_1 + a^m_{1j}) \Big] \label{eq:dp}
\end{equation}
\begin{equation}
m_j(z_1) = \max_{z_j} \Big[ \alpha_j \cdot \phi(I,z_j) + \beta_{1j} \cdot \psi(z_1 - z_j) \Big] \label{eq:dt}
\end{equation}
The maximization in Eqn.~\eqref{eq:dt} needs only be performed once across mixtures, and can be computed efficiently with a single min-convolution or distance transform~\citep{felzenszwalb2012distance}. The resulting message is then shifted by mixture-specific anchor positions $a^m_{1j}$ in Eqn.~\eqref{eq:dp}. Such mixture-independent messages can be computed only for leaf parts, because internal parts
in a tree will receive mixture-specific messages from downstream
children. Hence star EDPMs are essentially no more expensive than a EPM
(because a single min-convolution per part adds a negligible amount of
computation). In our experiments, running a 2000-mixture EDPM is almost as fast as a standard 6-mixture DPM. Other topologies beyond stars might provide greater flexibility. However, since EDPMs encode shape non-parametrically using many mixtures, each individual mixture may need not deform too much, making a
star-structured deformation model a reasonable approximation
(Fig.~\ref{fig:shape}).

\section{Experiments}
\label{sec:exp}
Armed with our array of non-parametric mixture models and datasets, we now present an extensive diagnostic analysis on 11 PASCAL categories from the 2010 PASCAL trainval set and faces from the Annotated Faces in the Wild test set \citep{zhu2012face}. For each category, we train the
model with varying number of samples ($N$) and mixtures ($K$).  To train our independent mixtures, we learn rigid HOG templates \citep{Dalal:CVPR05} with linear SVMs \citep{libsvm01a}. We calibrated SVM scores using Platt scaling~\citep{Platt99probabilisticoutputs}.
Since the goal is to calibrate scores of mixture components
relative to each other, we found it sufficient to train
scaling parameters using the original training set rather
than using a held-out validation set. To train our
compositional mixtures, we use a locally-modified variant of
the codebase from \citep{felzenszwalb2010object}. To show
the uncertainty of the performance with respect to different
sets of training samples, we randomly
re-sample the training data 5 times for each $N$ and $K$ following the
partitioned sampling scheme described in Sec.~\ref{sec:mixture}. The best
regularization parameter $C$ for the SVM was selected by cross validation. For
diagnostic analysis, we first focus on faces and buses.

{\bf Evaluation:}  We adopt the PASCAL VOC precision-recall
protocol for object detection (requiring 50\% overlap), and
report average precision (AP). While learning theory often
focuses on analyzing 0-1 classification error rather than
AP~\citep{mcallester1999some}, we experimentally verified
that AP typically tracks 0-1 classification error and so
focus on the former in our experiments.

\begin{figure}[t!]
\centering
(a) Single face template (test)\\
\includegraphics[trim=0mm 0mm 0mm 6mm, clip, width=0.7\columnwidth]{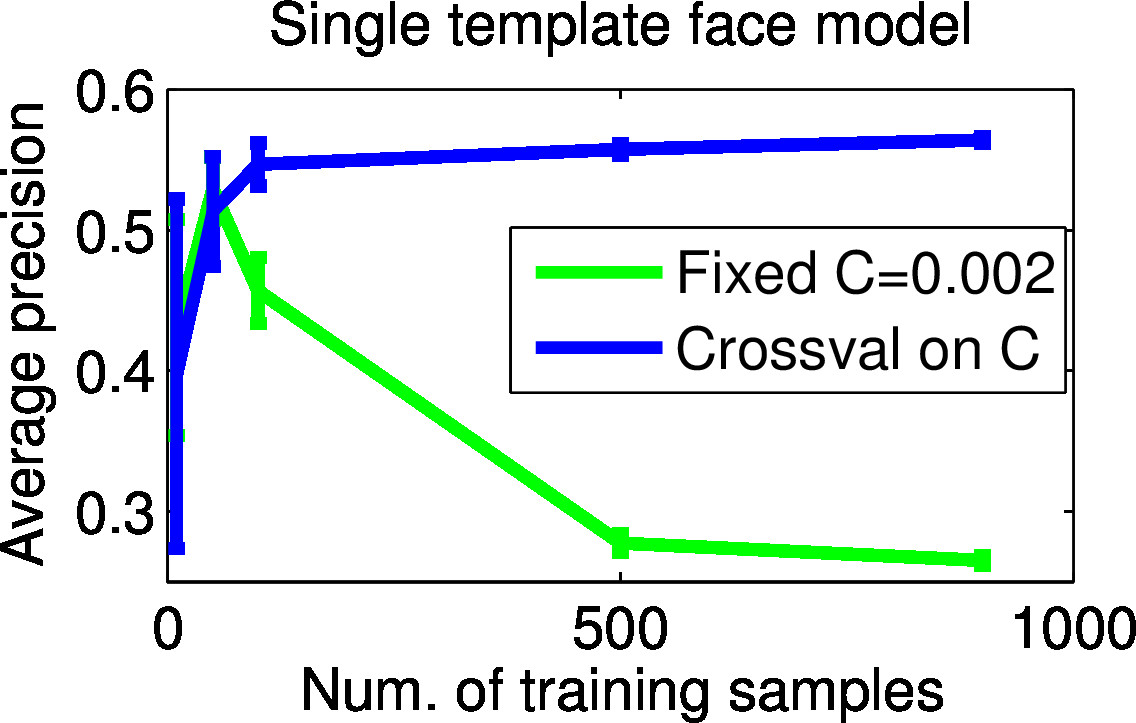}\\
\vspace{15pt}
(b) Single face template (train)\\
\includegraphics[width=0.7\columnwidth]{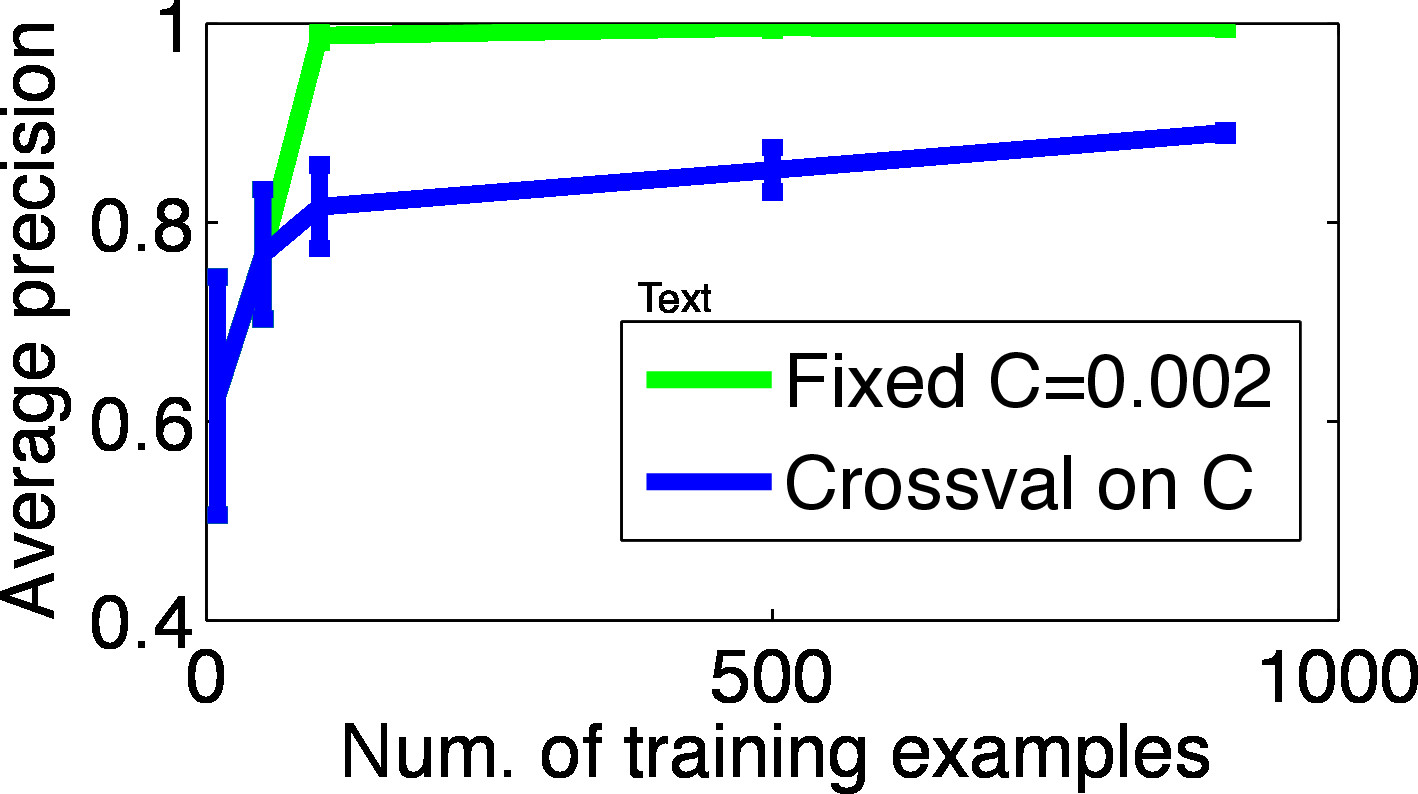}\\
\vspace{15pt}
(c) Single face template (test)\\
\includegraphics[trim=0mm 0mm 0mm 6mm, clip,width=0.7\columnwidth]{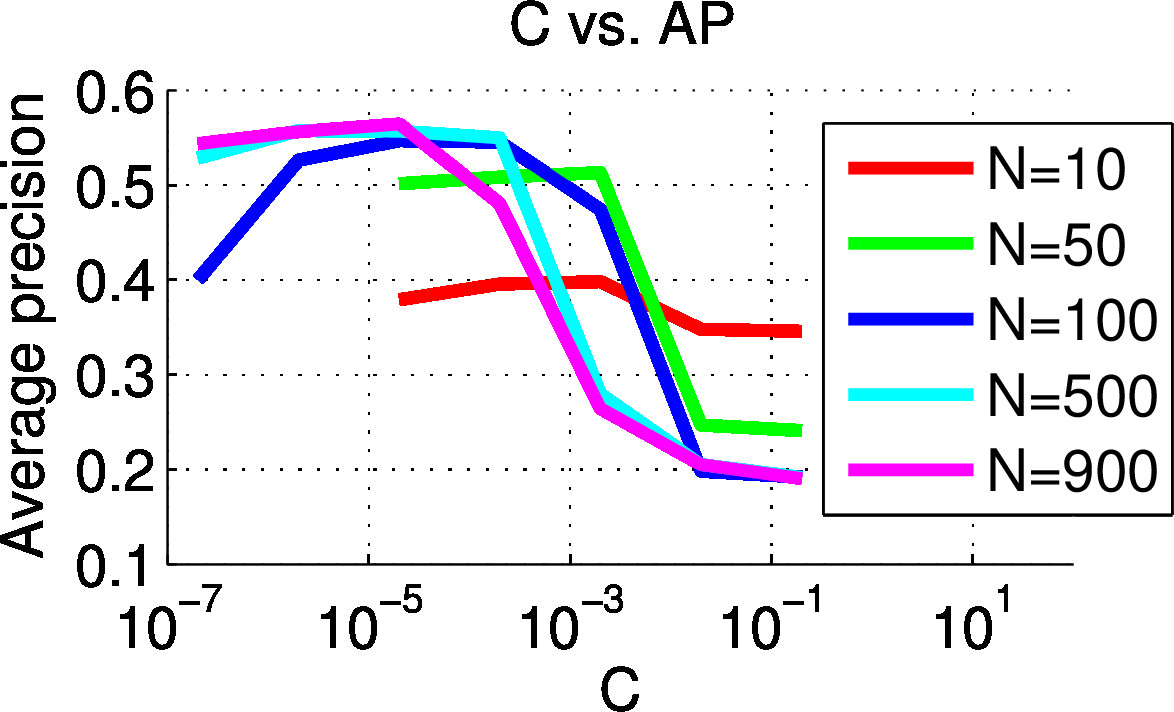}
\label{fig:tune_c}
\caption{ (a) More training data could hurt if we did not
cross-validate to select the optimal C. (b) Training error,
when measured on a fixed training set of 900 faces and 1218
negative images, always decreases as we train with more of
those images. This further suggests that overfitting is the
culprit, and that proper regularization is the solution.
(c) Test performance can change drastically with C.
Importantly, the optimal setting of C depends on the amount
of positive training examples $N$.}
\label{fig:facereg}
\end{figure}

\subsection{The importance of proper regularization} 

We begin with a rather simple experiment: how does a single rigid HOG template
tuned for faces perform when we give it more training data $N$?
Fig.~\ref{fig:facereg} shows the surprising result that additional training
data can \emph{decrease} performance!
For imbalanced object detection datasets with many more negatives than
positives, the hinge loss appears to grow linearly with the amount of positive
training data; if one doubles the number of positives, the total hinge loss
also doubles. This leads to overfitting. To address this problem, we found it
{\em crucial to cross-validate $C$} across different $N$. By doing so, we do
see better performance with more data (Fig.~\ref{fig:facereg}a).
While cross-validating regularization parameters is a standard procedure when applying a classifier
to a new dataset, most off-the-shelf detectors are trained using a fixed $C$
across object categories with large variations in the number of positives. We
suspect other systems based on standard detectors
\citep{felzenszwalb2010object,Dalal:CVPR05} may also be suffering from
suboptimal regularization and might show an improvement by proper
cross-validation.

\begin{figure}[t!]
\centering
\subfloat[]
{
\includegraphics[width=.55\columnwidth]{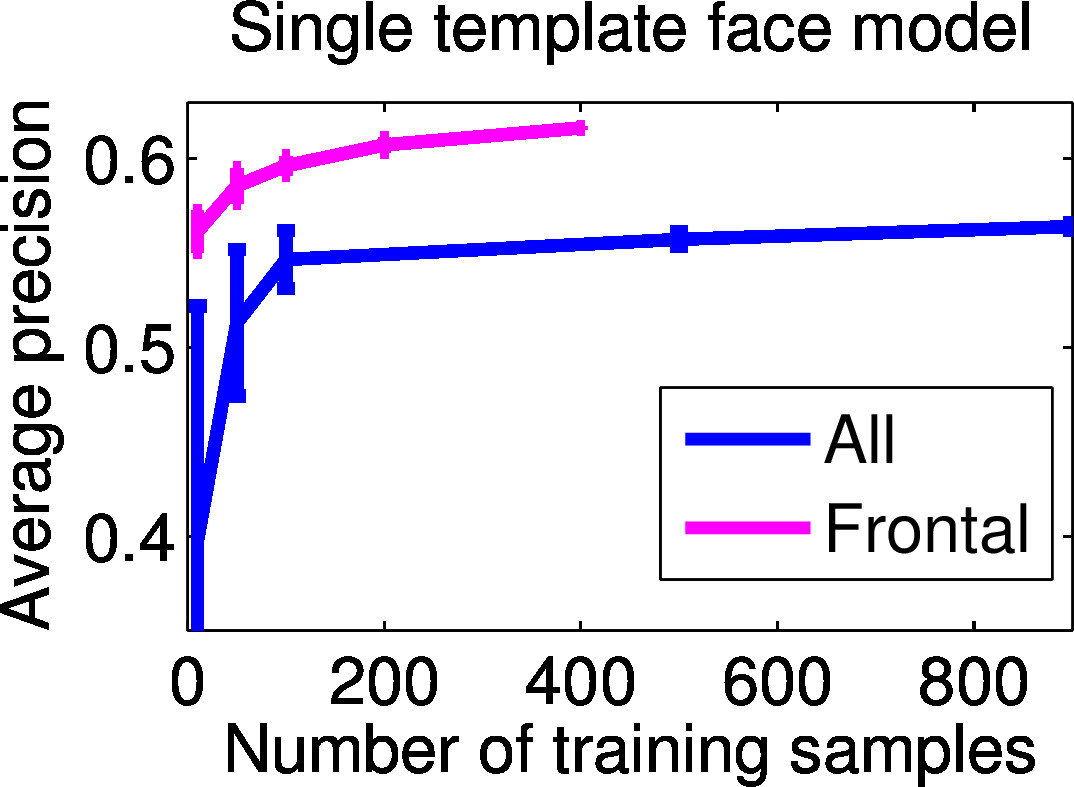}
} \\
\subfloat[Frontal]{
\includegraphics[width=.3\columnwidth]{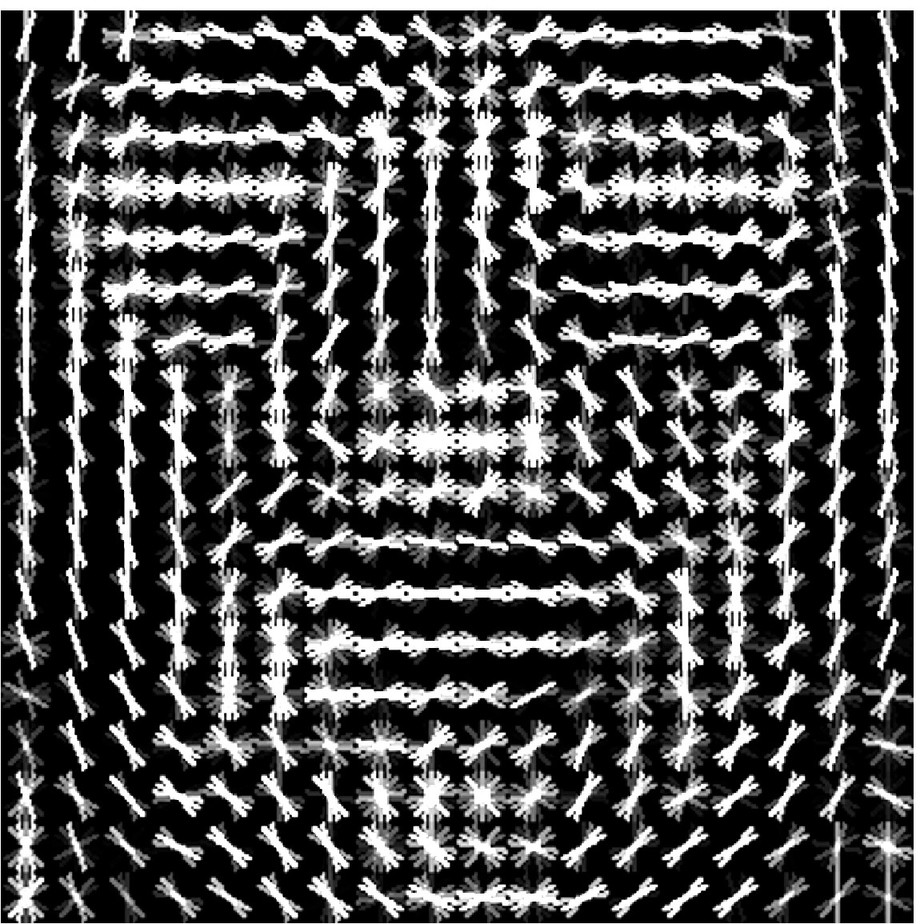}
}
\subfloat[All]{
\includegraphics[width=.3\columnwidth]{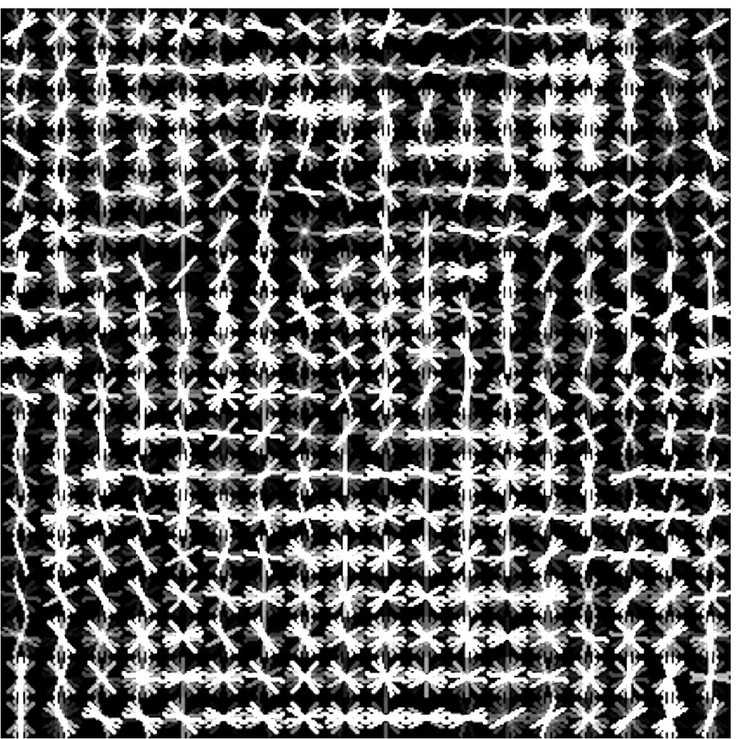}
}
\caption{In (a), we compare the performance of a single HOG template trained with
$N$ multi-view face examples, versus a template trained with a subset of those
$N$ examples corresponding to frontal faces. The
frontal-face template (b) looks ``cleaner'' and makes fewer
classification errors on both testing and training data. The
fully-trained template (c) looks noisy and performs worse,
even though it produces a lower SVM objective value (when
both (b) and (c) are evaluated on the full training set).
This suggests that SVMs are sensitive to noise and benefit
from training with ``clean'' data.}
\label{fig:faceclean}
\end{figure}

\begin{figure}[t!]
\centering
\includegraphics[width=\columnwidth]{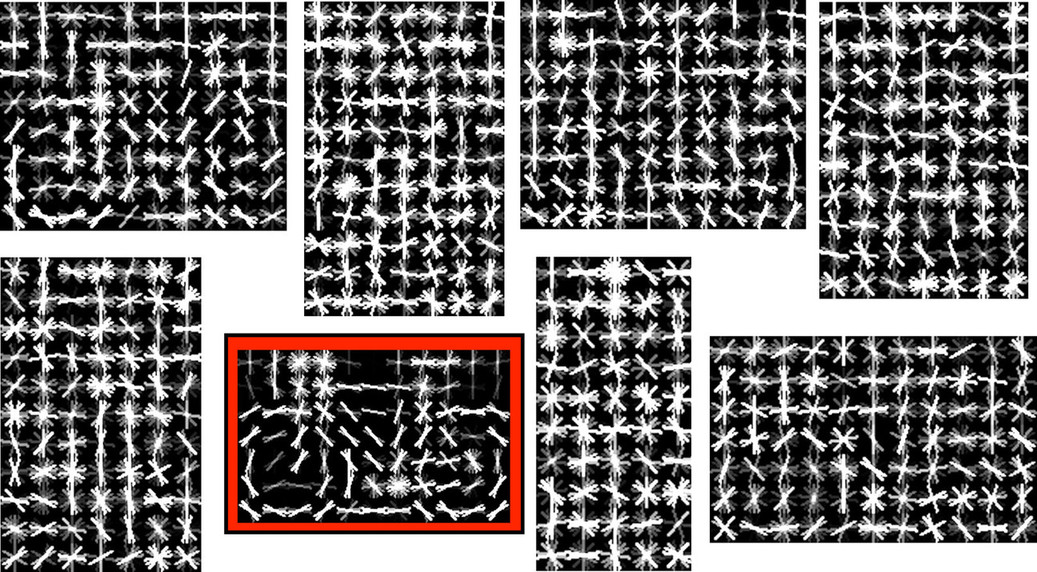}
\caption{The single bicycle template (marked with red) alone achieves
ap=29.4\%, which is almost equivalent to the performance of using all 8
mixtures (ap=29.7\%). Both models strongly outperform a single-mixture model
trained on the full training set. This suggests that these additional mixtures
are useful during training to capture outliers and prevent ``noisy'' data from
polluting a ``clean'' template that does most of the work at test time.}
\label{fig:bicycle_mix8}
\end{figure}

\subsection{The importance of clean training data}

Although proper regularization parameters proved to be crucial, we still
discovered scenarios where additional training data hurt performance.
Fig.~\ref{fig:faceclean} shows an experiment with a fixed set of $N$ training
examples where we train two detectors: (1) {\em All} is trained with with all
$N$ examples, while (2) {\em Frontal} is trained with a smaller, ``clean'' subset 
of examples containing frontal faces. We cross-validate $C$ for each model for each
$N$.  Surprisingly, {\em Frontal} outperforms {\em All} even though it is
trained with less data.

This outcome cannot be explained by a failure of the model to generalize from
training to test data. We examined the training loss for both models, evaluated
on the full training set. As expected, \emph{All} has a lower SVM objective
function than {\em Frontal} (1.29 vs 3.48).  But in terms of 0-1 loss, {\em
All} makes nearly twice as many classification errors on the same training
images (900 vs 470).  This observation suggests that {\em the hinge loss is a
poor surrogate to the 0-1 loss} because ``noisy'' hard examples can wildly
distort the decision boundary as they incur a large, unbounded hinge penalty.
Interestingly, latent mixture models can mimic the behavior of non-convex
bounded loss functions \citep{wu2007robust} by placing noisy examples into junk
clusters that simply serve to explain outliers in the training set.  In some
cases, a single ``clean'' mixture component by itself explains most of the test
performance (Fig.~\ref{fig:bicycle_mix8}).

\begin{figure}[t!]
\centering
Face\\
\includegraphics[width=0.6\columnwidth]{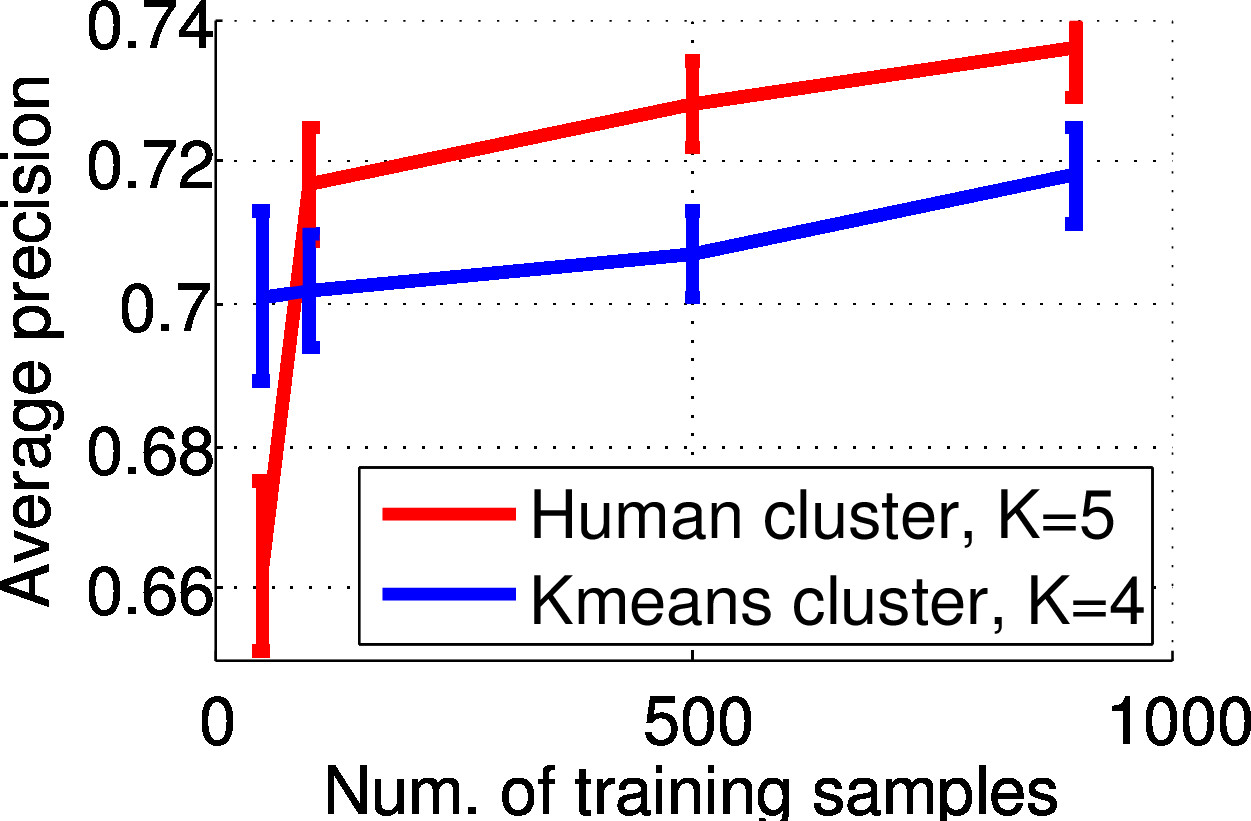}\\
\vspace{10pt}
Bus\\
\includegraphics[width=0.6\columnwidth]{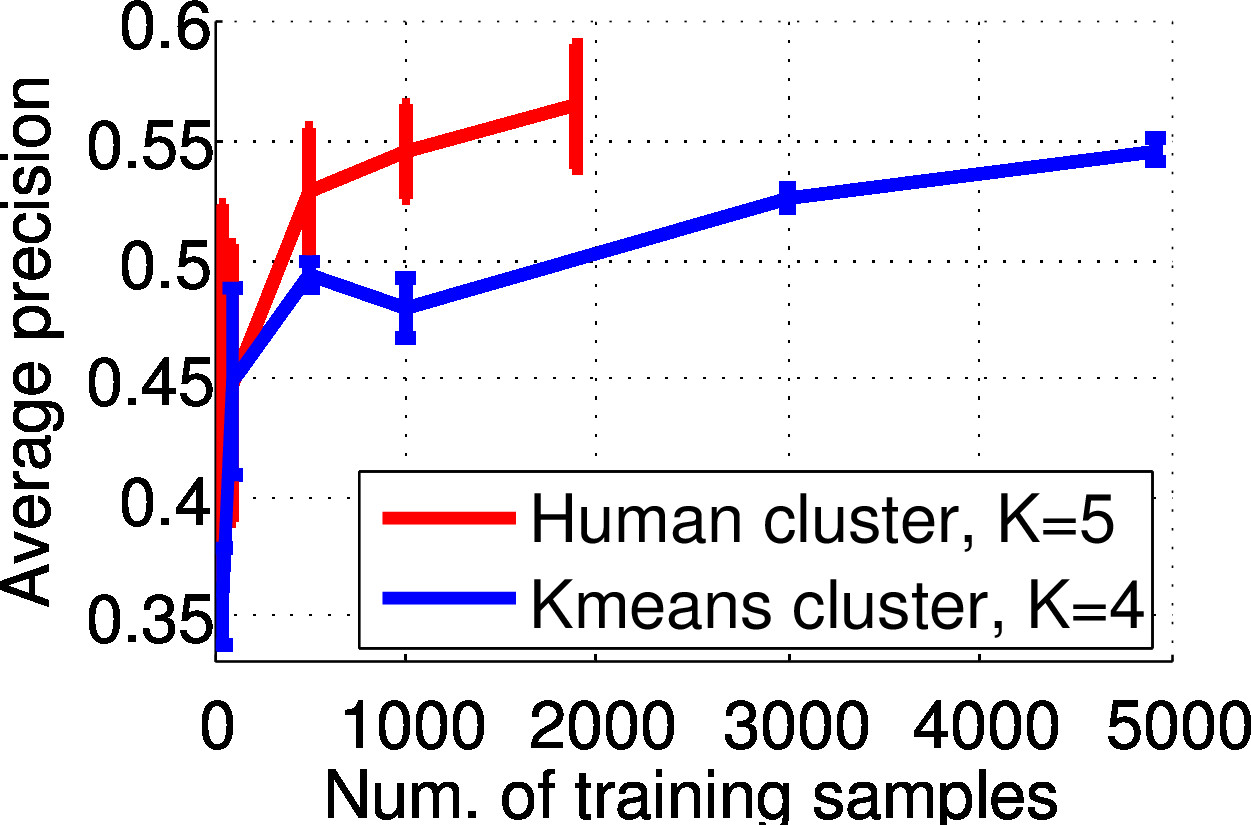}
\caption{We compare the human clustering and automatic k-means clustering at
near-identical $K$. We find that supervised clustering provides a small but
noticeable improvement of 2-5\%.}
\label{fig:clean}
\end{figure}

The importance of ``clean'' training data suggests it could be fruitful to
correctly cluster training data into mixture components where each component is
``clean''.  We evaluated the effectiveness of providing fully supervised
clustering in producing clean mixtures.  In Fig.~\ref{fig:clean}, we see a small
2\% to 5\% increase for manual clustering. In general, we find that
unsupervised clustering can work reasonably well but depends strongly on the
category and features used. For example, the DPM implementation of
\citep{felzenszwalb2010object} initializes mixtures based on aspect ratios.
Since faces in different viewpoint share similar aspect ratios, this tends to
produce ``unclean'' mixtures compared to our non-latent clustering.

\begin{figure}[t!]
\centering
\subfloat[Face (AP vs N)]
{
\hspace{-15pt}
\includegraphics[width=0.5\columnwidth,clip=true,trim=0mm 0mm 0mm 5mm]{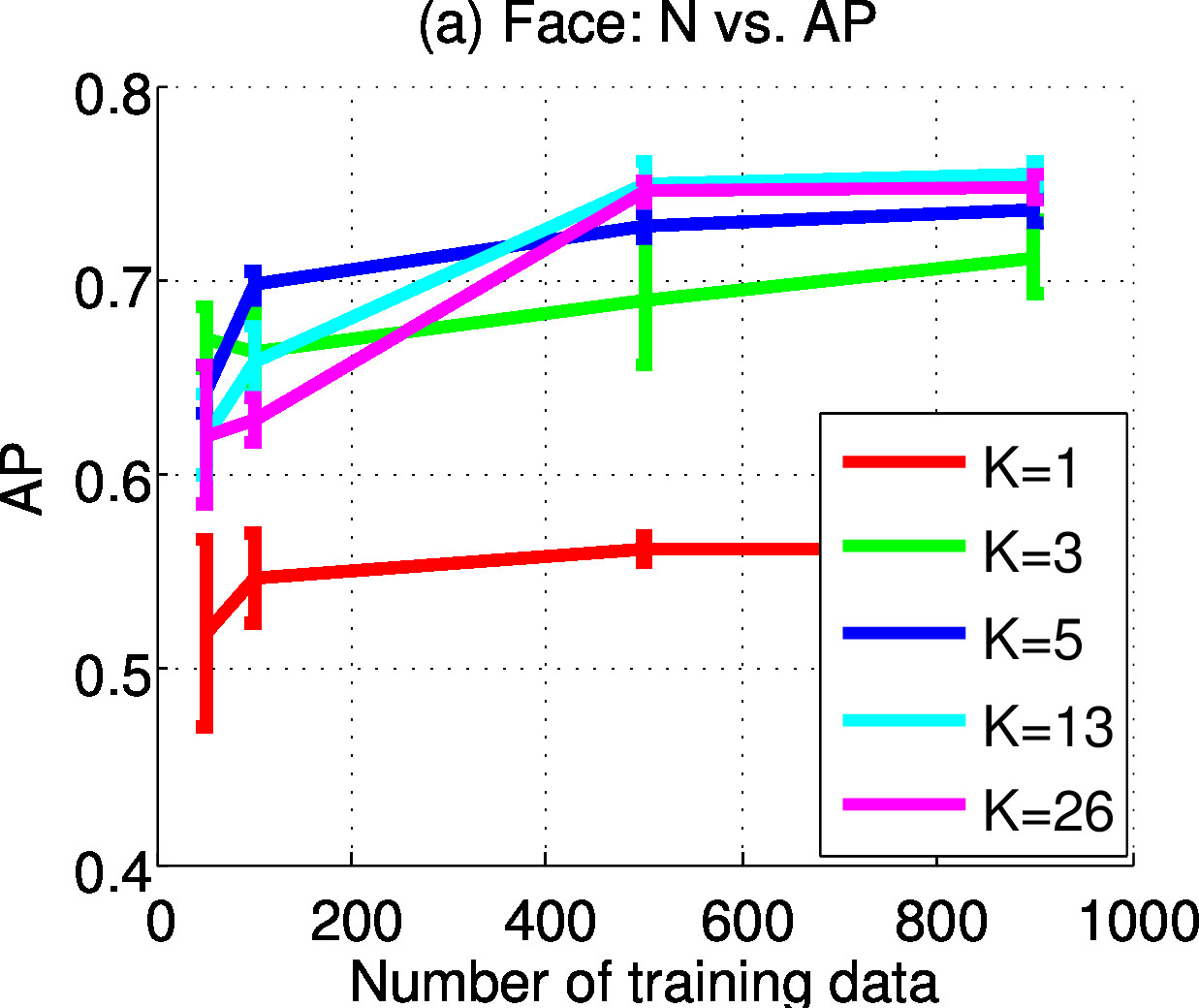}
\hspace{-15pt}
\label{fig:face_N}
} 
\subfloat[Face (AP vs K)]
{
\includegraphics[width=0.5\columnwidth,clip=true,trim=0mm 0mm 0mm 5mm]{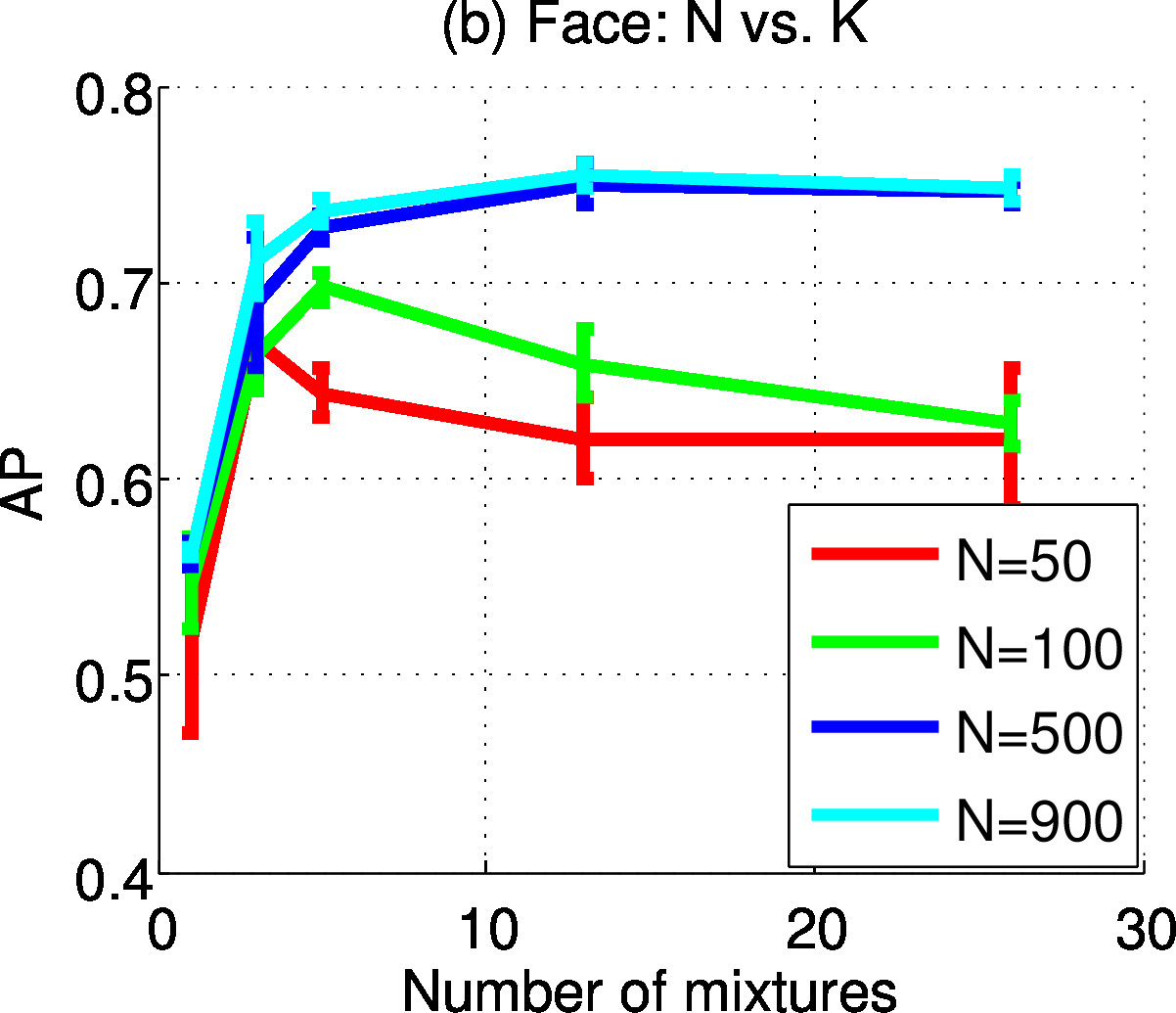}
\hspace{-15pt}
\label{fig:face_K}
} \\
\subfloat[Bus (AP vs N)]
{
\includegraphics[width=0.5\columnwidth,clip=true,trim=0mm 0mm 0mm 5mm]{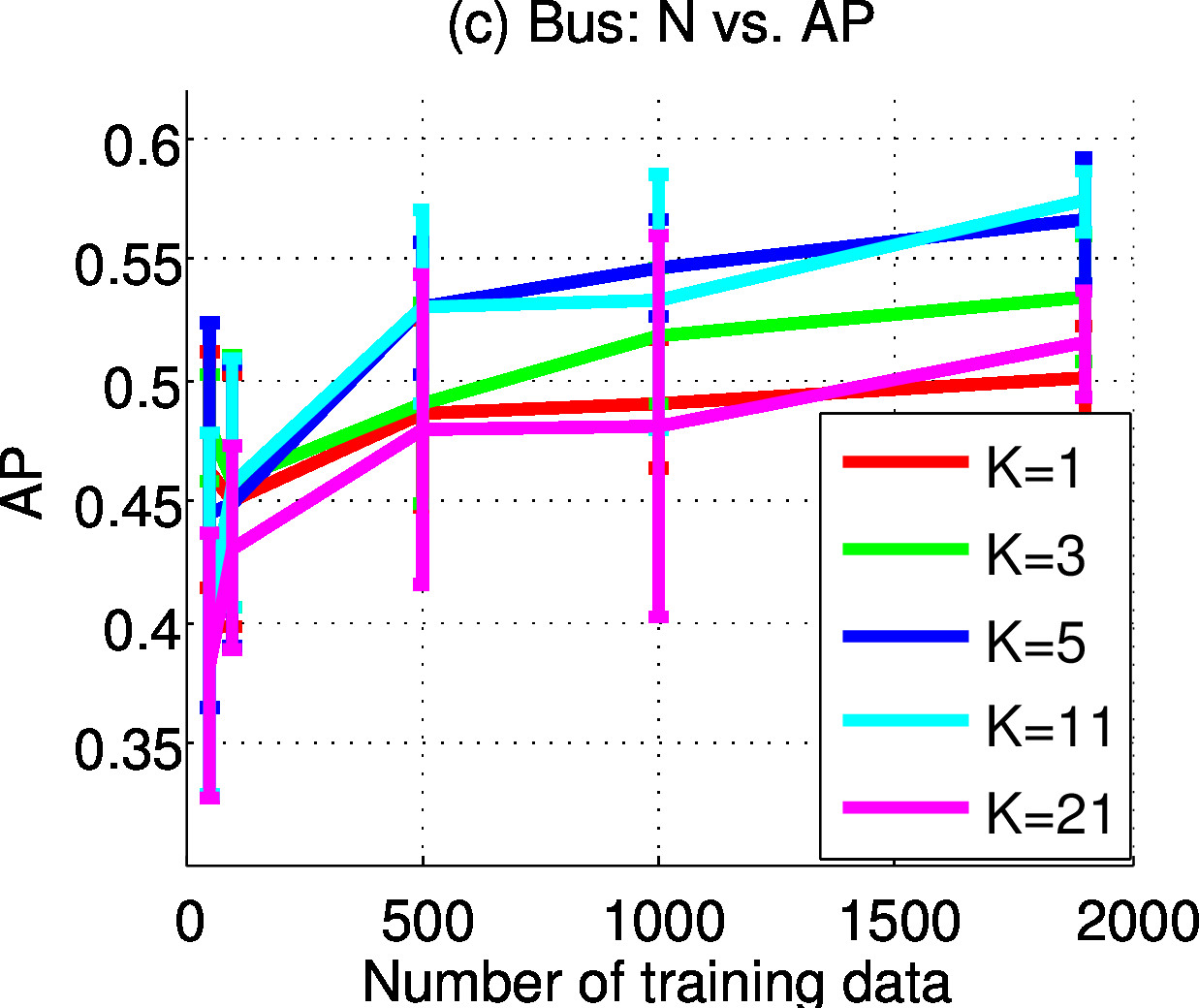}
\hspace{-15pt}
\label{fig:bus_N}
}
\subfloat[Bus (AP vs K)]
{
\includegraphics[width=0.5\columnwidth,clip=true,trim=0mm 0mm 0mm 5mm]{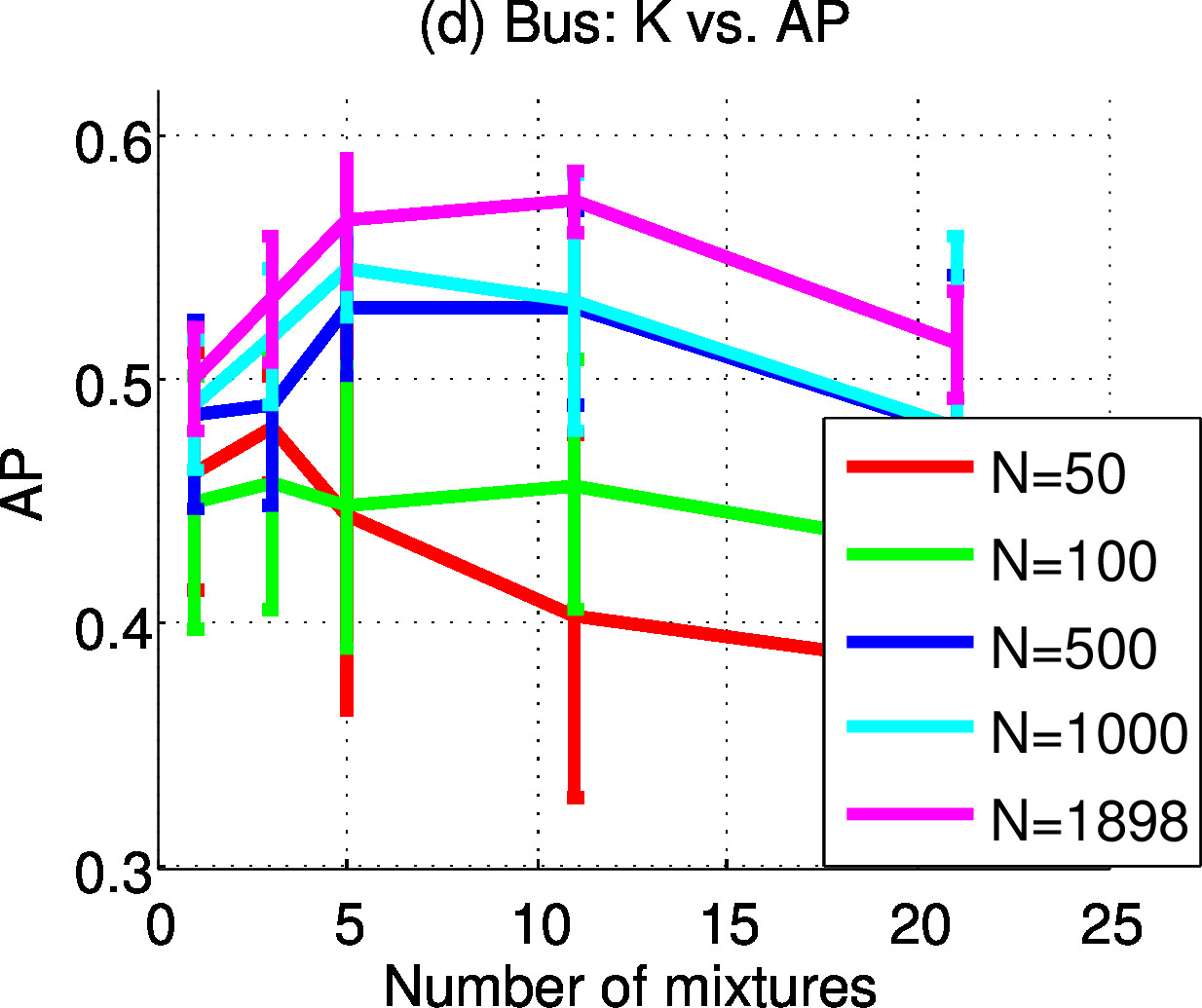}
\hspace{-15pt}
\label{fig:bus_K}
}
%\vspace{5pt}
\caption{(a)(c) show the monotonic non-decreasing curves when we add more
training data. The performance saturates quickly at a few hundred training
samples. (b)(d) show how the performance changes with more mixtures $K$. Given
a fixed number of training samples $N$, the performance increases at the
beginning, and decreases when we split the training data too much so that each
mixture only has few samples. }
\label{fig:face_bus_result}
\end{figure}

\subsection{Performance of independent mixtures}

Given the right regularization and clean mixtures trained independently, we now evaluate whether
performance asymptotes as the amount of training data and the model complexity
increase.

Fig.~\ref{fig:face_bus_result} shows
performance as we vary $K$ and $N$ after cross-validating $C$ and 
using supervised clustering. Fig.~\ref{fig:face_N} demonstrates that
increasing the amount of training data yields a clear improvement in performance
at the beginning, and the gain quickly becomes smaller later.
Larger models with more mixtures tend to perform worse with fewer examples due
to over fitting, but eventually win with more data. Surprisingly, improvement
tends to saturate at $\sim$100 training examples per
mixture and with $\sim$10
mixtures. Fig.~\ref{fig:face_K} shows performance as we vary model
complexity for a fixed amount of training data. Particularly at small data
regimes, we see the critical point one would expect from Fig.~\ref{fig:splash}:
a more complex model performs better up to a point, after which it overfits. We
found similar behavior for the buses category which we manually clustered by
viewpoint.

\begin{figure}
\begin{center}
\subfloat[]
{
\includegraphics[width=1\columnwidth]{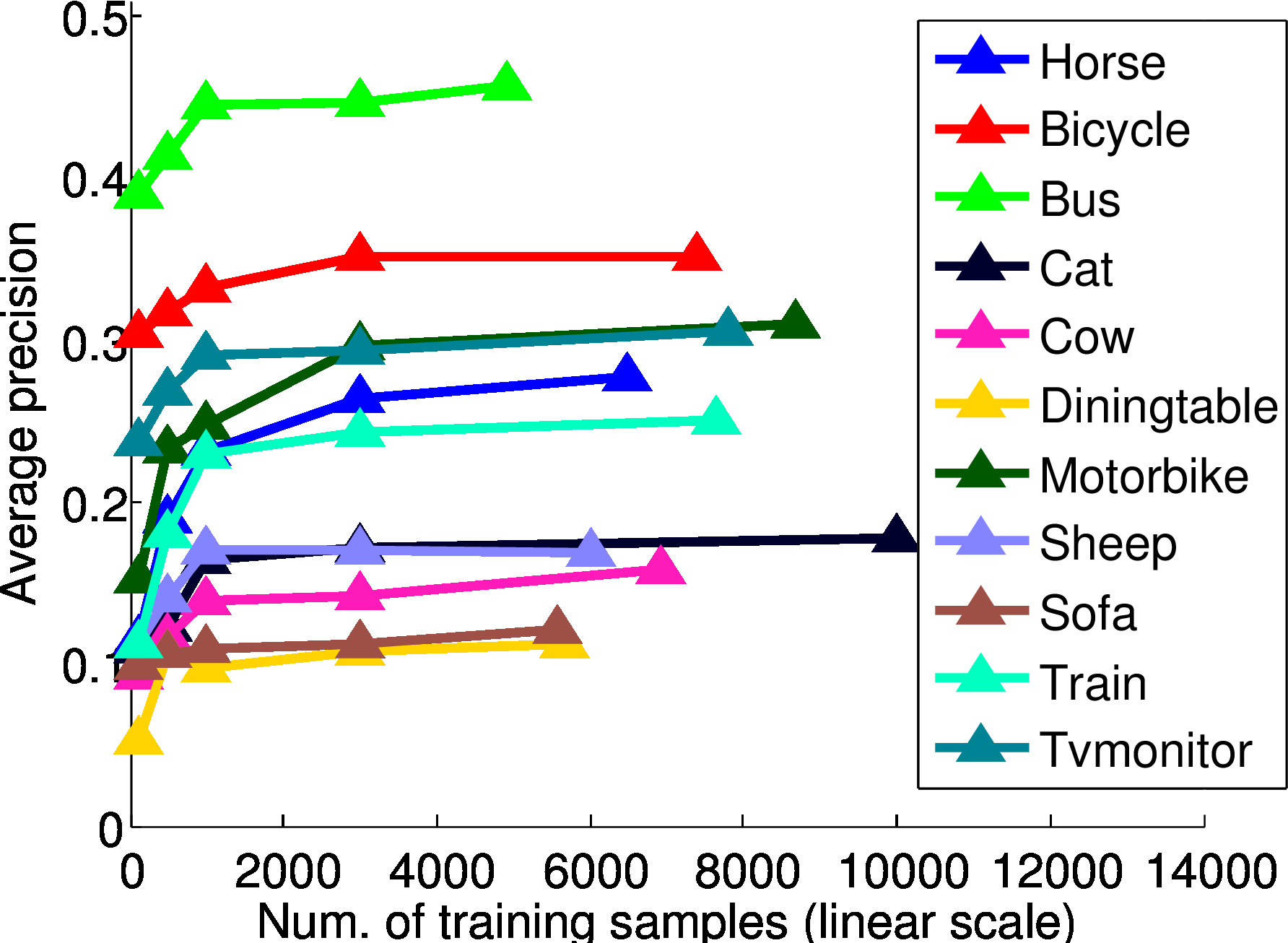}
\label{pascal_linear}
} \\
\subfloat[]
{
\includegraphics[width=1\columnwidth]{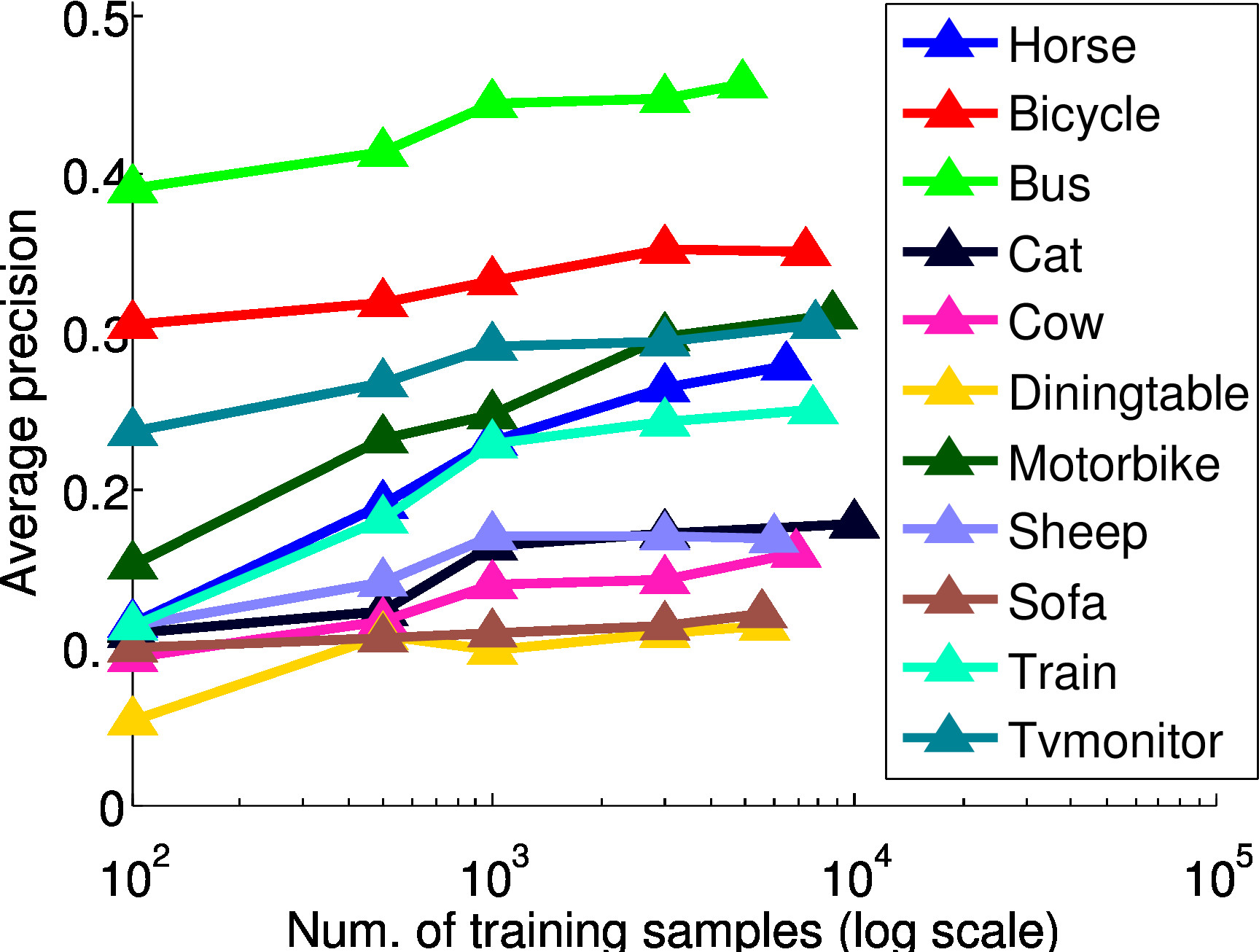}
\label{pascal_log}
}
\caption{We plot the best performance at varying amount of training data for
11 PASCAL categories on PASCAL 2010 trainval set. (a) shows that all the curves
look saturated with a relatively small amount of training data; but in log scale (b)
suggests a diminishing return instead of true saturation. However the
performance increases so slow that we will need more than $10^{12}$ examples per
category to
reach 95\% AP if we keep growing at the same rate.}
\label{fig:pascal}
\end{center}
\end{figure}

We performed similar experiments for all 11 PASCAL object categories in our
PASCAL-10X dataset shown in Fig.~\ref{fig:pascal}. We evaluate
performance on the PASCAL 2010 trainval set since the testset annotations are
not public. We cluster the training data into $K$=[1,2,4,8,16]
mixture components, and $N$=[50, 100, 500, 1000, 3000, $N_{max}$] training
samples, where $N_{max}$ is the number of training samples collected for 
the given category.  For each $N$, we
select the best $C$ and $K$ through cross-validation. Fig.~\ref{pascal_linear},
appears to suggest that performance is saturating across all categories as we
increase the amount of training data. However, if we plot performance on a log
scale (Fig.~\ref{pascal_log}), it appears to increase roughly linearly. This
suggests that the required training data may need to grow exponentially to
produce a fixed improvement in accuracy. For example, if we extrapolate the
steepest curve in Fig.~\ref{pascal_log} (motorbike), we will need {\bf
$10^{12}$} motorbike samples to reach 95\% AP!

Of course 95\% AP may not be an achievable level of performance. There is some
upper-bound imposed by the Bayes risk associated with the HOG feature space
which no amount of training data will let us surpass.  {\em Are classic mixtures of rigid templates approaching the Bayes optimal performance?}  Of course we cannot compute the Bayes risk so this is hard to answer in general.  However, the performance of any system operating on the same data and feature space provides a lower bound on the optimal performance. We next analyze the performance of compositional mixtures to provide much better lower bound on optimal performance.

\subsection{Performance of compositional mixtures}
\begin{figure}[t!]
\begin{center}
Faces
\includegraphics[width=\columnwidth]{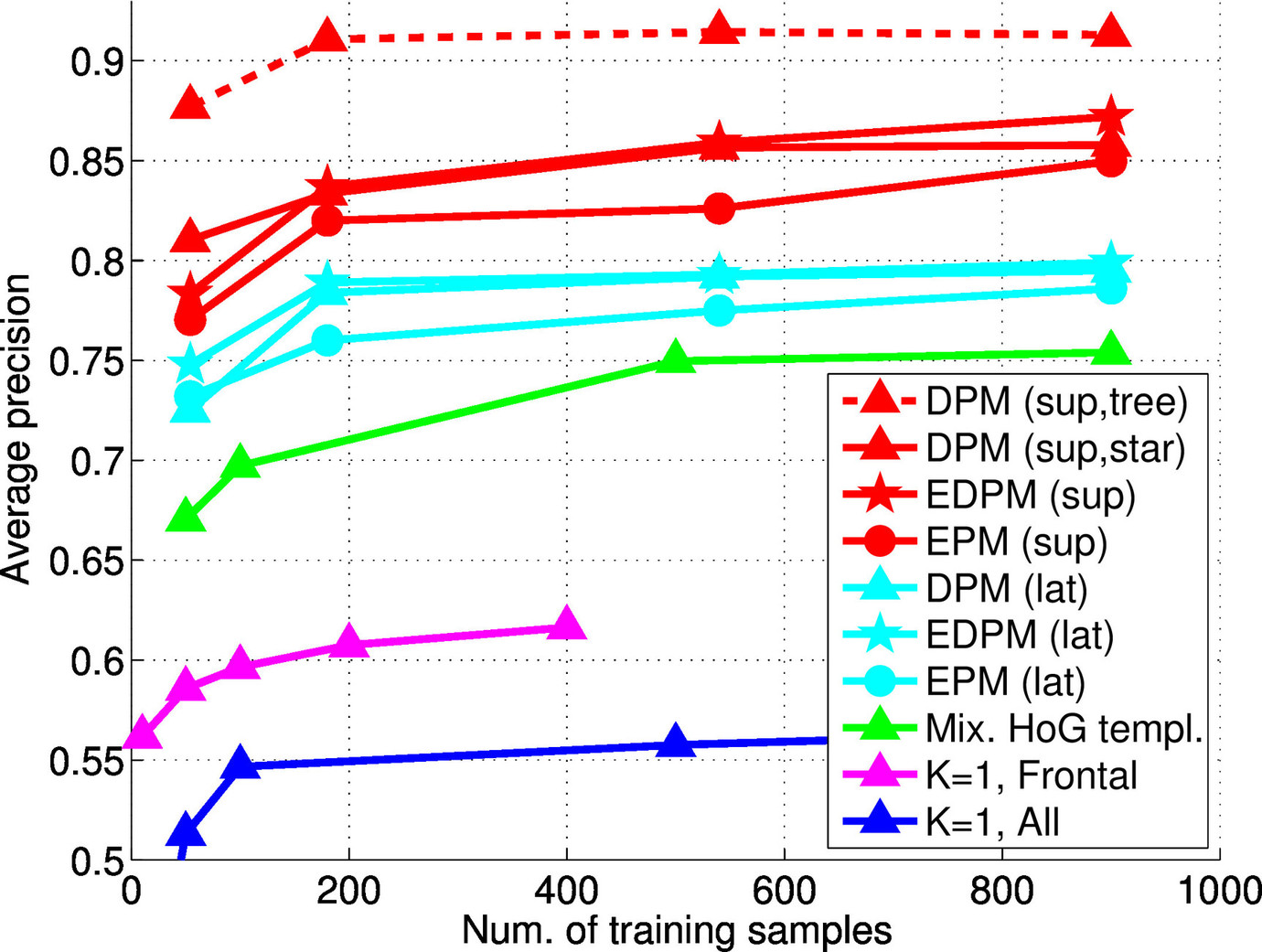}
\caption{We compare the performance of mixtures models with EPMs
  and latent/supervised DPMs for the task of face detection. A single rigid template ($K=1$) tuned for
  frontal faces outperforms the one tuned for all faces (as shown in
  Fig.~\ref{fig:faceclean}). Mixture models boost performance to 76\%,
  approaching the performance of a latent DPM (79\%). The EPM shares supervised
  part parameters across rigid templates, boosting performance to 85\%.
  The supervised DPM (91\%) shares parameters but also implicitly scores
  additional templates not seen during training.}
\label{fig:bayes}
\end{center}
\end{figure}

\begin{figure}[t!]
\begin{center}
Bus\\
\includegraphics[width=0.9\columnwidth]{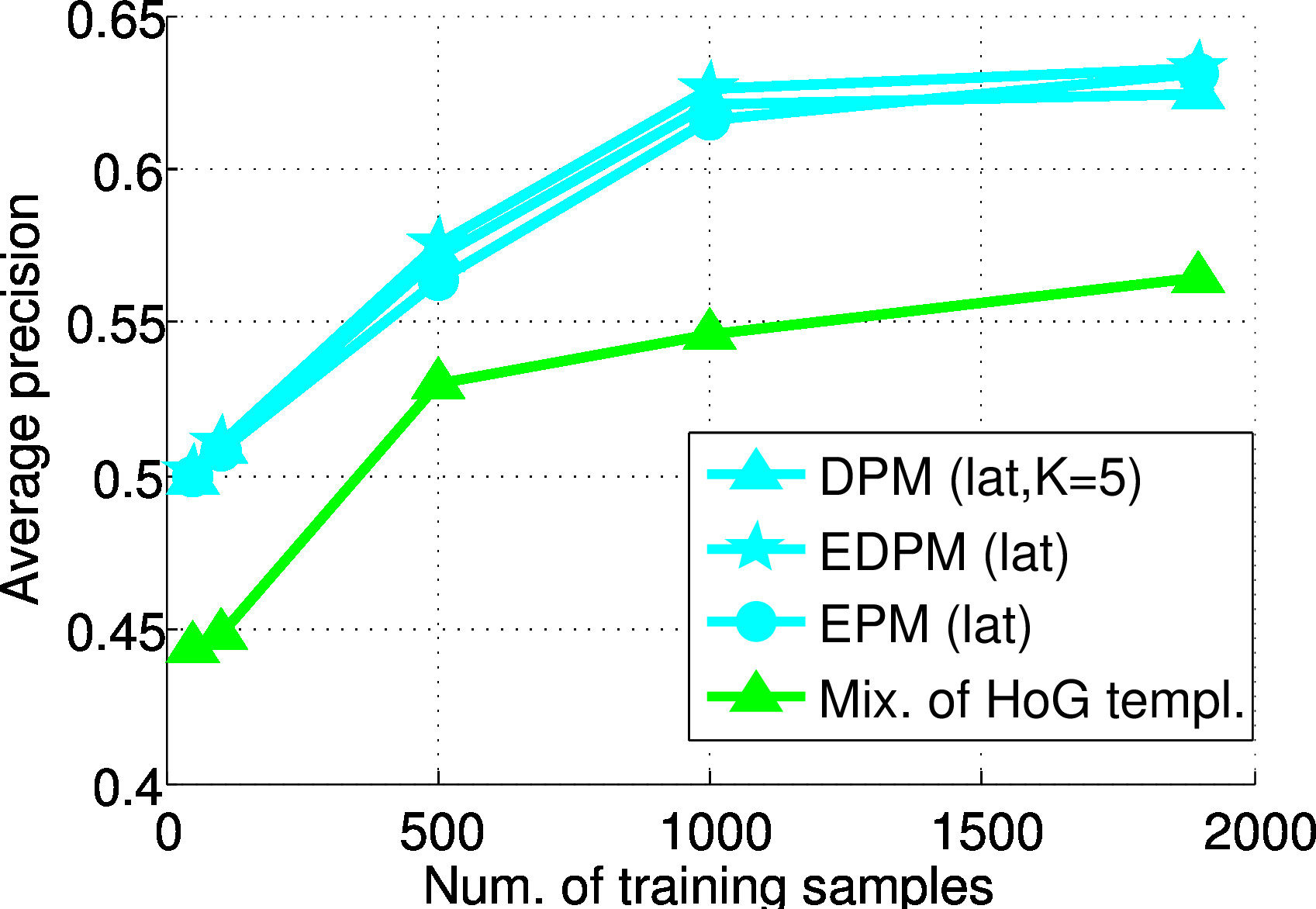}
\caption{We compare the performance of mixture models with latent EPMs, EDPMs, and DPMs for bus detection. In the latent setting, EPMs significantly outperform the rigid mixtures of template and match the performance of the standard latent DPMs.}
\label{fig:bus_bayes}
\end{center}
\end{figure}

We now perform a detailed analysis of compositional mixture models, including DPMs, EPMs, and EDPMs. We focus on face detection and Pascal buses. We consider the latent star-structured DPM of \citep{felzenszwalb2010object} as our primary baseline. For face detection, we also compare to the supervised tree-structured DPM of \citep{zhu2012face}, which uses facial landmark annotations in training images as supervised part locations. Each of these DPMs makes use of different parts, and so can be used to define different EPMs and EDPMs. We plot performance of faces in Fig.~\ref{fig:bayes} and buses in Fig.\ref{fig:bus_bayes}.

{\bf Supervised DPMs:} For face detection, we first note that a supervised DPM can perform quite well (91\% AP) with less than 200 example faces. This represents a lower bound on the maximum achievable performance with a mixture of linear templates given a fixed training set. This performance is noticeably higher than that of our cross-validated rigid mixture model, which maxes out at an AP of 76\% with 900 training examples. By extrapolation, we predict that one would need $N=10^{10}$ training examples to achieve the DPM performance. To analyze where this performance gap is coming from, we now evaluate the performance of various compositional mixtures models.

{\bf Latent parts:} We begin by analyzing the performance of compositional
mixtures defined by latent parts, as they can be constructed for both faces and
Pascal buses. Recall that EPMs have the benefit of sharing parameters between
rigid templates, but they cannot extrapolate to new shape configurations not
seen among the $N$ training examples. EPMs noticeably improve performance over
independent mixtures, improving AP from 76\% to 78.5\% for faces and improving
AP from 56\% to 64\% for buses. In fact, for large $N$, they approach the
performance of latent DPMs, which is 79\% for faces and 63\% for buses. For
small $N$, EPMs somewhat underperform DPMs. This makes sense: with very few
observed shape configurations, exemplar-based methods are limited. But
interestingly, with a modest number of observed shapes ($\approx$ 1000),
exemplar-based methods with parameter sharing can approach the performance of
DPMs. This in turn suggests that extrapolation to unseen shapes is may not be
crucial, at least in the latent case. This is further evidenced by the fact that
EDPMs, the deformable counterpart to EPMs, perform similarly to both EPMs and DPMs.

{\bf Supervised parts:} The story changes somewhat for supervised
parts. Here, supervised EPMs outperform independent mixtures 85\% to
76\%. Perhaps surprisingly, EPMs even outperform latent DPMs. However,
supervised EPMs still underperform a supervised DPM. This suggests that, in the
supervised case, {\em the performance gap (85\% vs 91\%) stems from
  the ability of DPMs to synthesize configurations that are not seen
  during training}. Moreover, the reduction in relative error due to
extrapolation is more significant than the reduction due to part
sharing. \citep{zhu2012face} point out that a tree-structured DPM
significantly outperforms a star-structured DPM, even when both are
trained with the same supervised parts. One argument is that trees
better capture nature spatial constraints of the model, such as the contour-like continuity of small parts. Indeed, we also find that a star-structured DPM does a ``poorer'' job of extrapolation. In fact, we show that an EDPM does as well a supervised star model, but not quite up to the performance of a tree DPM.

\begin{figure}[t!]
\begin{center}
\subfloat[Bus]{
\includegraphics[trim=0mm 6mm 0mm 0mm, clip, width=0.8\columnwidth]{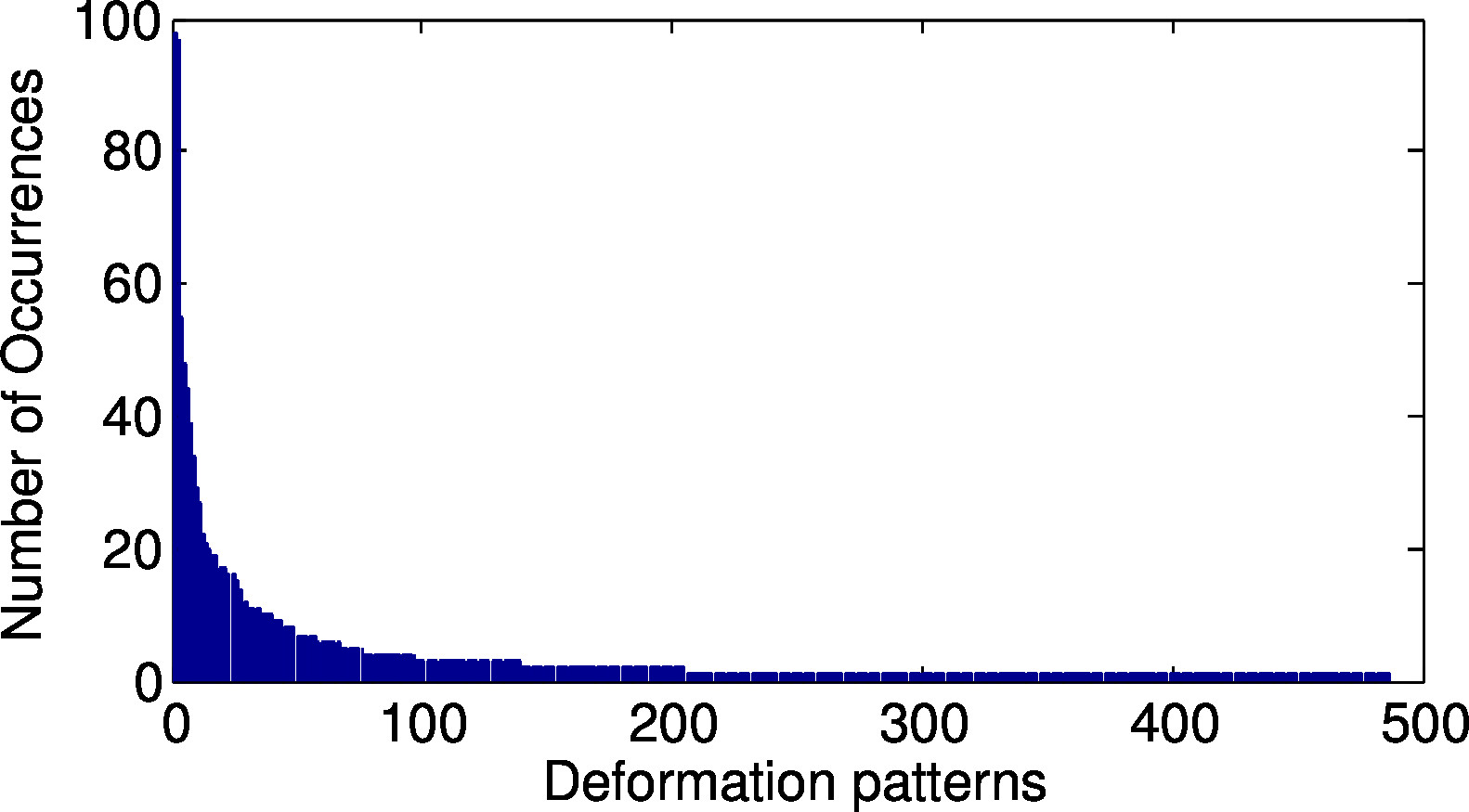}
} \\
\subfloat[Face]{
\includegraphics[trim=0mm 6mm 0mm 0mm, clip,width=0.8\columnwidth]{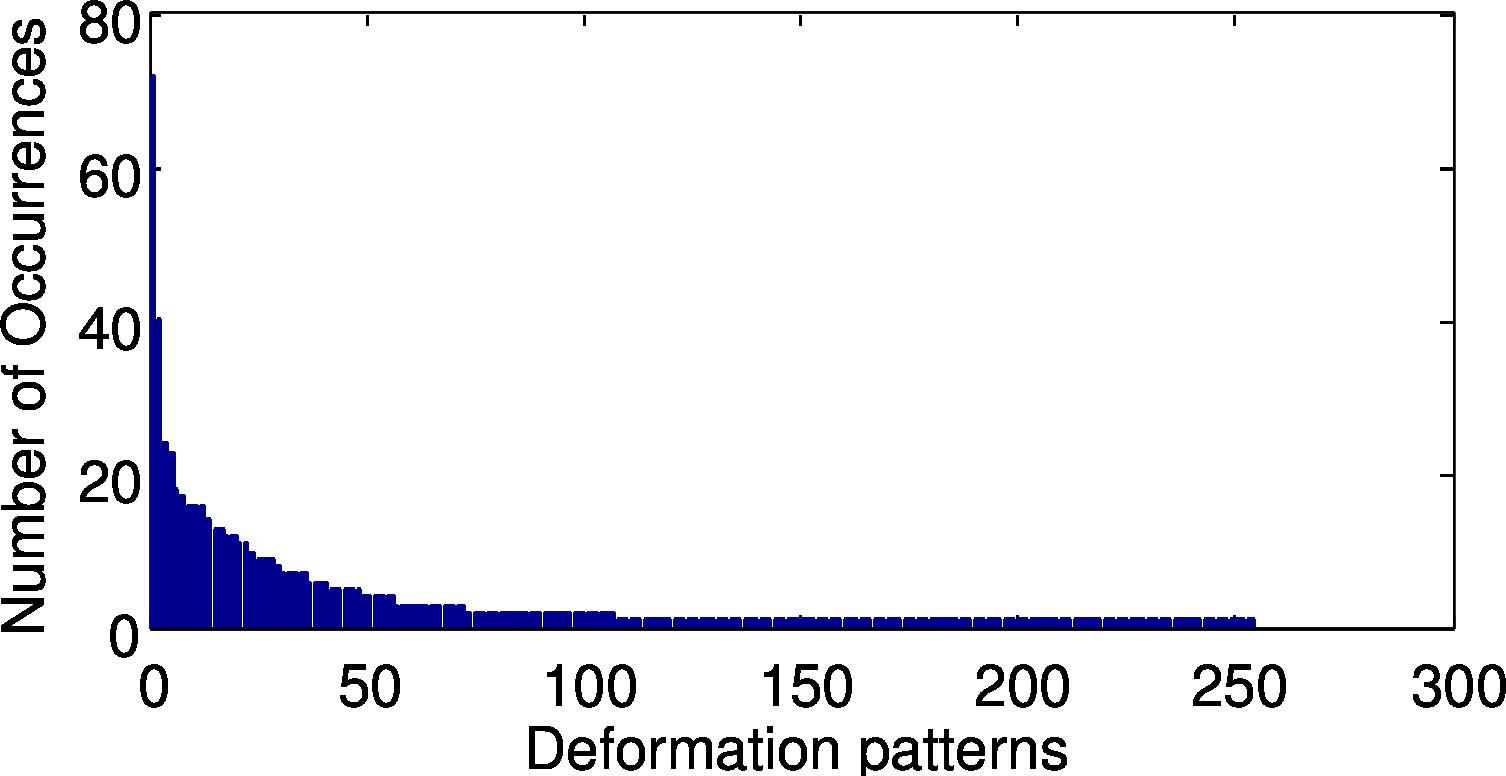}
}
\caption{We plot the number of distinct shape patterns in
our training set of buses and faces. Each training example
is ``binned'' into a discrete shape by quantizing a vector
of part locations. The above histograms count the number of
examples that fall into a particular shape bin.  In both
cases, the number of occurrences seems to follow a long-tail
distribution: a small number of patterns are common, while
there are a huge number of rare cases. Interestingly, there
are less than 500 unique bus configurations observed in our
PASCAL-10X dataset of 2000 training examples.  This suggests
that one can build an exemplar part model (EPM) from the
``right'' set of 500 training examples and still perform
similarly to a DPM trained on the full dataset
(Fig.~\ref{fig:bus_bayes}).}
\label{fig:bus_longtail}
\end{center}
\end{figure}

{\bf Analysis:} Our results suggest that part models can be seen as a mechanism for performing intelligent
parameter sharing across observed mixture components and extrapolation to
implicit, unseen mixture components. Both these aspects
contribute to the strong performance of DPMs. However, with
the ``right'' set of (supervised) parts and the ``right''
geometric (tree- structured) constraints, extrapolation to
unseen templates has the potential to be much more
significant. We see this as a consequence of the
``long-tail'' distribution of object shape
(Fig.~\ref{fig:bus_longtail}); many object instances can be
modeled with a few shape configurations, but there exists of
long tail of unusual shapes. Examples from the long tail may
be difficult to observe in any finite training dataset,
suggesting that extrapolation is crucial for recognizing
these cases.  Once the representation for sharing and
extrapolation is accurately specified, fairly little
training data is needed. Indeed, our analysis shows that one
can train a state-of-the-art face detector
\citep{zhu2012face} with 50 face images.

\begin{figure}[t!]
\begin{center}
\includegraphics[width=\columnwidth]{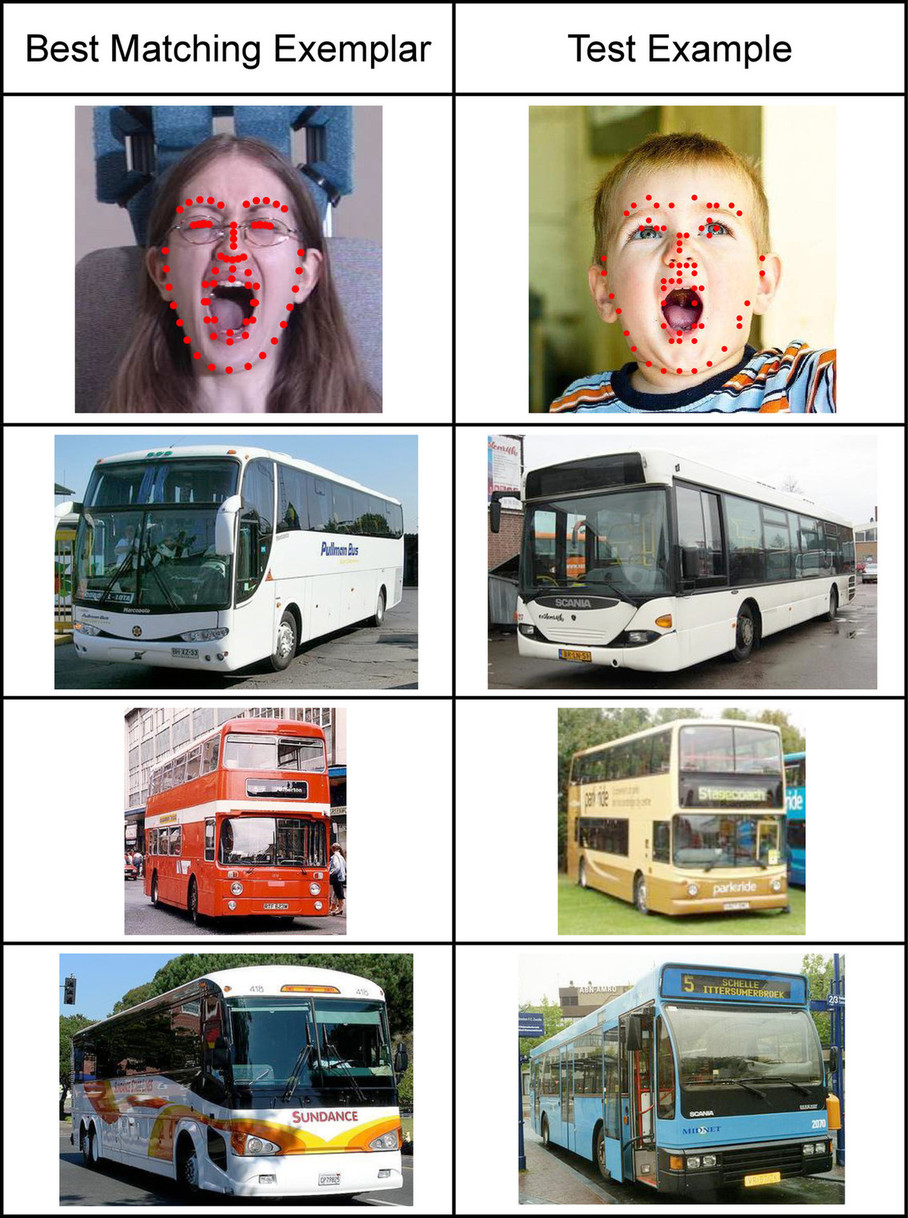}
\caption{We visualize detections using our exemplar DPM
(EDPM) model. As opposed to existing exemplar-based methods
\citep{malisiewicz2011ensemble}, our model shared parameters
between exemplars (and so is faster to evaluate) and can
generalize to unseen shape configurations. Moreover, EDPMs
returns corresponding landmarks between an exemplar and a
detected instance (and hence an associated set of landmark
deformation vectors), visualized on the top row of faces.}
\label{fig:label_transfer}
\end{center}
\end{figure}

{\bf Relation to Exemplar SVMs:} In the setting of object detection,
we were not able to see significant performance improvements due to our
non-parametric compositional mixtures. However, EDPMs may be useful for other
tasks. Specifically, they share an attractive property of exemplar SVMs (ESVMs)
\citep{malisiewicz2011ensemble}: each detection can be affiliated with its
closest matching training example (given by the mixture index), allowing us to
transfer annotations from a training example to the test instance.
\citep{malisiewicz2011ensemble} argue that non-parametric label transfer is an
effective way of transferring associative knowledge, such as 3D pose,
segmentation masks, attribute labels, etc. However, unlike ESVMs, EDPMs share
computation among the exemplars (and so are faster), can generalize to unseen
configurations (since they can extrapolate to new shapes), and also report a
part deformation field associated with each detection (which maybe useful to
warp training labels to better match the detected instance). We show example
detections (and their matching exemplars) in Fig.~\ref{fig:label_transfer}.

\section{Related Work}
\label{sec:conclusion}
We view our study as complementary to other meta-analysis of the object recognition problem, such as
studies of the dependence of performance on the number of object categories
\citep{deng2010}, visual properties \citep{hoiem2012eccv},
dataset collection bias \citep{torralba2011unbiased}, and
component-specific analysis of recognition pipelines \citep{parikh2011finding}.

{\bf Object detection:} Our analysis is focused on template-based
approaches to recognition, as such methods are currently competitive
on challenging recognition problems such as PASCAL. However, it
behooves us to recognize the large body of alternate approaches
including hierarchical or ``deep'' feature learning \citep{krizhevsky2012imagenet},
local feature analysis \citep{tuytelaars2008local}, kernel methods
\citep{vedaldi2009multiple}, and decision
trees~\citep{bosch2007image}, to name a few. Such methods may produce
different dependencies on performance as a function of dataset size
due to inherent differences in model architectures. We hypothesize
that our conclusions regarding parameter sharing and extrapolation may
generally hold for other architectures.

{\bf Non-parametric models in vision:} Most relevant to our analysis is work on data-driven models for recognition. Non-parametric scene models have been used for scene
completion~\citep{hays2007scene}, geolocation~\citep{hays2008im2gps}. Exemplar-based methods have also been used for scene-labeling through label transfer~\citep{liu2011nonparametric,tighe2010superparsing}. Other examples include nearest-neighbor methods for low-resolution image
analysis~\citep{torralba200880} and image
classification~\citep{zhang2006svm,boiman2008defense}. The closest
approach to us is \citep{malisiewicz2011ensemble}, who learn exemplar
templates for object detection. Our analysis suggests that it is crucial to
share information between exemplars and extrapolate to unseen templates by
re-composing parts to new configurations.

{\bf Scalable nearest-neighbors:} We demonstrate that compositional part models are
one method for efficient nearest-neighbor computations. Prior work has
explored approximate methods such as hashing
\citep{shakhnarovich2003fast,shakhnarovich2005nearest} and kd-trees
\citep{muja2009fast,beis1997shape}. Our analysis shows that one can view
parts as tools for exact and efficient indexing into an exponentially-large set
of templates. This suggests an alternative perspective of parts as
computational entities rather than semantic ones.

\section{Conclusion}
We have performed an extensive analysis of the current
dominant paradigm for object detection using HOG feature
templates. We specifically focused on performance as a
function of the amount of training data, and introduced
several non-parametric models to diagnose the state of
affairs. 

To scale current systems to larger datasets, we find that one must get certain ``details'' correct. Specifically, (a) cross-validation of
regularization parameters is mundane but crucial, (b) current
discriminative classification machinery is overly sensitive to noisy data,
suggesting that (c) manual cleanup and supervision or more clever latent
optimization during learning may play an important role for designing
high-performance detection systems. We also demonstrate that HOG templates have
a relatively small effective capacity; one can train accurate HOG templates with 100-200 positive examples (rather than thousands of examples as is typically done \citep{Dalal:CVPR05}).

From a broader perspective, an emerging idea in our community is that object detection might be solved with
simple models backed with massive training sets. Our experiments suggest
a slightly refined view.
Given the size of existing datasets, it appears that the current
state-of-the-art will need significant additional data (perhaps exponentially
larger sets) to continue producing consistent improvements in performance. We
found that larger gains were possible by enforcing richer constraints within
the model, often through non-parametric compositional representations that could
make better use of additional data.  In some sense, we need ``better models'' to
make better use of ``big data''.

Another common hypothesis is that we should focus on developing better
features, not better learning algorithms. While HOG is certainly limited, we
still see substantial performance gains without any change in the features
themselves or the class of discriminant functions. Instead, the strategic
issues appear to be parameter sharing, compositionality, and non-parametric
encodings. Establishing and using accurate, clean correspondence among training
examples (e.g., that specify that certain examples belong to the same
sub-category, or that certain spatial regions correspond to the same part) and
developing non-parametric compositional approaches that implicitly make use of
augmented training sets appear the most promising directions.

\bibliographystyle{spbasic}      % basic style, author-year citations
\bibliography{references}   % name your BibTeX data base

\end{document}